\documentclass[letterpaper, 10 pt, conference]{ieeeconf}  % Comment this line out if you need a4paper

\IEEEoverridecommandlockouts                              % This command is only needed if 
\input{paper_style.sty}

\title{\LARGE \bf
Using Clarke Transform to Create a Framework on the Manifold:\\
From Sampling via Trajectory Generation to Control
}

\author{Reinhard M.~Grassmann and Jessica Burgner-Kahrs% <-this % stops a space
\thanks{We acknowledge the support of the Natural Sciences and Engineering Research Council of Canada (NSERC), [RGPIN-2019-04846].}
\thanks{All authors are with Continuum Robotics Laboratory, Department of Mathematical and Computational Sciences, University of Toronto, Mississauga, ON L5L 1C6, Canada {\tt\small reinhard.grassmann@utoronto.ca}}
}

\begin{document}

\maketitle
\thispagestyle{empty}
\pagestyle{empty}

\begin{abstract}

We present a framework based on Clarke coordinates for spatial displacement-actuated continuum robots with an arbitrary number of joints.
This framework consists of three modular components, \textit{i.e.}, a planner, trajectory generator, and controller defined on the manifold.
All components are computationally efficient, compact, and branchless, and an encoder can be used to interface existing framework components that are not based on Clarke coordinates.
We derive the relationship between the kinematic constraints in the joint space and on the manifold to generate smooth trajectories on the manifold.
Furthermore, we establish the connection between the displacement constraint and parallel curves.
To demonstrate its effectiveness, a demonstration in simulation for a displacement-actuated continuum robot with four segments is presented.

\end{abstract}

\section{Introduction}

Soft and continuum robots must necessarily move from a current configuration to a desired configuration to accomplish meaningful real-world tasks in medical and industrial applications.
Thus, a minimal framework should consist of three parts; planner, trajectory generator, and controller.
However, developed frameworks are tailored to a specific robot morphology rendering some, if not all, components of the framework unsuitable for other robot morphologies.
This limits the reusability and rapid advancement in the field.
Overcoming that by providing a modular framework, where each component has the same interface, has obvious benefits, \textit{e.g.},
(i) interpretability and safety due to a divide-and-conquer approach,
(ii) compatibility due to the same interface,
(iii) flexibility in changing and testing components,
(vi) agility in developing, improving, and testing of separate components, and
(v) empowering the field by sharing and reusing components.
Overall, an increase in cost-efficiency and a decrease in the number of times the wheel needs to be reinvented can be expected.
Therefore, providing a general principle to create frameworks is a necessary step to move forward together, rather than just improving one particular robot morphology.

As a candidate for an interface between the components, different improved state representations \cite{DellaSantinaBicchiRus_RAL_2020, AllenAlbert_et_al_RoboSoft_2020, CaoXie_et_al_JMR_2022, DianGuo_et_al_Access_2022, GrassmannSenykBurgner-Kahrs_arXiv_2024} have been introduced in recent years.
It has been shown that using improved state representations has decisive advantages.
Della Santina \textit{et al.} \cite{DellaSantinaBicchiRus_RAL_2020} discuss and analyze the main issues arising from the use of commonly used parameterization, \textit{i.e.}, curvature $\kappa$ and bending plane angle $\theta$ used for constant curvature models.
The derived improved state representation \cite{DellaSantinaBicchiRus_RAL_2020} for $n = 4$ number of joints shows superb performance on a model-based control application.
Allen \textit{et al.} \cite{AllenAlbert_et_al_RoboSoft_2020} introduces improved state representations for $n = 3$ and $n = 4$ number of joints.
They propose closed-form and singularity-free forward kinematics as well as Jacobian-based inverse kinematics mitigating the disadvantages of previous approaches.
Dian \textit{et al.} \cite{DianGuo_et_al_Access_2022} proposed an improved state representation that can be seen as an adaptation of improved state representation by Della Santina \textit{et al.} \cite{DellaSantinaBicchiRus_RAL_2020} to $n = 3$.
As mentioned by Dupont \textit{et al.} \cite{DupontRucker_et_al_JPROC_2022}, different representations exist and among them, that utilize Cartesian curvature components, \textit{i.e.}, $\kappa_x$ and $\kappa_y$, have the advantage of avoiding a parametric singularity.
The connection between the Cartesian curvature components and all improved state representations is established by Grassmann \textit{et al.} \cite{GrassmannSenykBurgner-Kahrs_arXiv_2024}.

Grassmann \textit{et al.} \cite{GrassmannSenykBurgner-Kahrs_arXiv_2024} links the displacement-actuated joints and the displacement constraint to electrical currents and Kirchhoff's current law, respectively.
This and the reformulation of the joint representation lead to the use of the Clarke transformation matrix that allows the disentanglement of the displacement constraint. 
As a result, an improved state representation called Clarke coordinates is derived unifying previous improved state representations \cite{DellaSantinaBicchiRus_RAL_2020, AllenAlbert_et_al_RoboSoft_2020, CaoXie_et_al_JMR_2022, DianGuo_et_al_Access_2022, GrassmannSenykBurgner-Kahrs_arXiv_2024} and generalizing them to $n \geq 3$.
Since constant curvature \cite{WebsterJones_IJRR_2010} is not enforced during their derivation, the Clarke coordinates are not limited to constant curvature assumption.
Grassmann \& Burgner-Kahrs \cite{GrassmannBurgner-Kahrs_arXiv_2024} show that the analogy to the Clarke transformation matrix is not necessary to derive the so-called Clarke transform.
This allows for consideration of arbitrary joint location distribution on the constant cross-section of a displacement-actuated continuum robot.

In general, the derived approaches based on improved state representations are closed-from, compact, computationally efficient, and singularity-free.
This makes the Clarke coordinates a desirable representation and unified language to interface all components within a framework.
Due to the recent introduction of the Clarke transform, only sampling methods \cite{GrassmannSenykBurgner-Kahrs_arXiv_2024, GrassmannBurgner-Kahrs_arXiv_2024} and controllers \cite{GrassmannSenykBurgner-Kahrs_arXiv_2024, GrassmannBurgner-Kahrs_arXiv_2024} have been proposed.
While previous sampling methods and controllers can be used as components in a framework, a trajectory generator has yet to be formulated using Clarke coordinates.

This work considers a displacement-actuated continuum robot defined by kinematic design parameters consisting of segment length $l = \textit{const.}$, distance $d_i$, angle $\psi_i$, and number of joints $n$.
We kindly refer to \cite{GrassmannSenykBurgner-Kahrs_arXiv_2024, GrassmannBurgner-Kahrs_arXiv_2024} for a depiction.
For $n$ equally distributed joints, the polar coordinate given by
\begin{align}
    \psi_i = \dfrac{2\pi}{n} (i - 1)
    \quad\text{and}\quad
    d_i = d = \textit{const.}
    \label{eq:psi}
\end{align}
describes the location of the $i\textsuperscript{th}$ joint in the cross-section.
The kinematic design parameters fully describe the displacement-actuated continuum robot in the kinematic sense encompassing a large class of soft robots and continuum robots.

In the presented work, we aim to shift the paradigm towards a unified approach by introducing a framework, where each of the components is based on the Clarke coordinates of a \SI{2}{dof} manifold.
In particular, the following contributions are made:
\begin{itemize}
    \item Presentation of a framework completely based on the Clarke coordinates 
    \item Introduction of a smooth trajectory generator that specifies and generates trajectories on the \SI{2}{dof} manifold
    \item Derivation of the kinematics constraints for trajectories on the manifold
    \item Connection between displacement constraint and parallel curves also known as offset curves
\end{itemize}
\section{Clarke Transform}

In this section, we provide a self-contained and brief description of the Clarke transform, the relationship to arc parameters, and virtual displacement.
The description of virtual displacement is expanded to the connection between the displacement constraint and parallel curves.
Furthermore, we briefly state the encoder-decoder architecture.
For a longer discussion, we kindly refer to \cite{GrassmannSenykBurgner-Kahrs_arXiv_2024, GrassmannBurgner-Kahrs_arXiv_2024, FirdausVadali_AIR_2023}.

\subsection{Clarke Transform for Displacement-Actuated Joint}

Proposed by Grassmann \textit{at al.} \cite{GrassmannSenykBurgner-Kahrs_arXiv_2024}, a vector with values of each displacement-actuated joint $\rho_i$ can be represented as
\begin{align}
    \rhovec = 
    \begin{bmatrix} 
        \rho_1 \\ \rho_2 \\ \vdots \\ \rho_n 
    \end{bmatrix}
    =
	\begin{bmatrix}
		\rhoreal\cos\left(\psi_1\right) + \rhoim\sin\left(\psi_1\right) \\
        \rhoreal\cos\left(\psi_2\right) + \rhoim\sin\left(\psi_2\right) \\
        \vdots \\
        \rhoreal\cos\left(\psi_n\right) + \rhoim\sin\left(\psi_n\right)
	\end{bmatrix}
    \in
    \mathbf{Q}
    ,
	\label{eq:rho}
\end{align}
where $\mathbf{Q} \subset \mathbb{R}^n$ is the joint space.
All joint values in \eqref{eq:rho} are interdependent and constrained obeying the displacement constraint given by
\begin{align}
	\sum_{i=1}^{n} \rho_{i} = 0.
    \label{eq:sum_rho}
\end{align}
The Clarke coordinates \cite{GrassmannSenykBurgner-Kahrs_arXiv_2024}, \textit{i.e.}, $\rhoreal$ and $\rhoim$ in \eqref{eq:rho}, can be combined into a vector as well, \textit{i.e.},
\begin{align}
    \rhoclarke =    
	\begin{bmatrix}
        \rhoreal \\
        \rhoim 
	\end{bmatrix}
    .
    \label{eq:rho_clarke}
\end{align}
Both representations can be transformed into each other via
\begin{align}
    \rhoclarke &= \MP\rhovec
    \quad\text{and}\quad
    \label{eq:forward}
    \\
    \rhovec &= \MPinv\rhoclarke,
    \label{eq:inverse}
\end{align}
where the generalized Clarke transformation matrix $\MP$ is
\begin{align}
	\MP
    =
    \dfrac{2}{n}\!
	\begin{bmatrix}
		\cos\left(\psi_0\right) & \cos\left(\psi_1\right) & \cdots & \cos\left(\psi_n\right)\\%[1em]
		\sin\left(\psi_0\right) & \sin\left(\psi_1\right) & \cdots & \sin\left(\psi_n\right)
	\end{bmatrix}
    ,
	\label{eq:MP}
\end{align}
which holds for symmetrically arranged joints \cite{GrassmannSenykBurgner-Kahrs_arXiv_2024, GrassmannBurgner-Kahrs_arXiv_2024}.
By factor out \eqref{eq:rho_clarke} in \eqref{eq:rho}, its inverse $\MPinv$ can be found being
\begin{align}
	\MPinv
    =
	\begin{bmatrix}
		\cos\left(\psi_1\right) & \sin\left(\psi_1\right) \\%[1em]
        \cos\left(\psi_2\right) & \sin\left(\psi_2\right) \\%[1em]
        \vdots & \vdots\\%[1em]
        \cos\left(\psi_n\right) & \sin\left(\psi_n\right)
	\end{bmatrix}
    ,
	\label{eq:MP_inverse}
\end{align}
where \eqref{eq:MP_inverse} is the right-inverse of non-square matrix \eqref{eq:MP}.

\subsection{Arc Parameters}

Utilizing the constant curvature assumption \cite{WebsterJones_IJRR_2010}, it is possible to identify the arc parameters, \textit{i.e.}, curvature $\kappa$ and bending-plane angle $\theta$.
Derived by Grassmann \textit{et al.} \cite{GrassmannSenykBurgner-Kahrs_arXiv_2024}, this relation between Clarke coordinates and arc parameters is
\begin{align}
    \rhoclarke
    =
    \begin{bmatrix}
        \rhoreal \\ 
        \rhoim
    \end{bmatrix}
    =
    \begin{bmatrix}
        dl\kappa\cos\left(\theta\right) \\ 
        dl\kappa\sin\left(\theta\right)
    \end{bmatrix}
    .
    \label{eq:relation_to_arc}
\end{align}
Note that $l$ is constant and not considered as arc parameters in this context.
As hinted by the used analogy and naming convention by Grassmann \textit{et al.} \cite{GrassmannSenykBurgner-Kahrs_arXiv_2024}, we can tread \eqref{eq:relation_to_arc} as a complex number embedded in $\mathbb{R}^2$.
The modulus and argument of \eqref{eq:relation_to_arc} is given by
\begin{align}
    |\rhoclarke|
    &=
    dl\kappa
    \quad\text{and}
    \label{eq:rho_clarke_modulus}
    \\
    \arg\rhoclarke
    &=
    \theta.
    \label{eq:rho_clarke_argument}
\end{align}
As can be seen, \eqref{eq:rho_clarke_modulus} is the scaled curvature $\kappa$ or the scaled bending angle $\phi = l\kappa$, whereas \eqref{eq:rho_clarke_argument} is bending-plane angle $\theta$.
Other arguments and {formul\ae} are presented in \cite{GrassmannSenykBurgner-Kahrs_arXiv_2024, GrassmannBurgner-Kahrs_ICRA_EA_2024}.
To further investigate the geometric meaning of $|\rhoclarke|$, we take a look at the concept of virtual displacement \cite{FirdausVadali_AIR_2023, GrassmannSenykBurgner-Kahrs_arXiv_2024} and the displacement constraint \eqref{eq:sum_rho}.

\subsection{Virtual Displacement and Displacement Constraint}
\label{sec:virtual_displacement}

\begin{figure}
    \centering
    \vspace*{0.75em}
    \includegraphics[width=0.4\columnwidth]{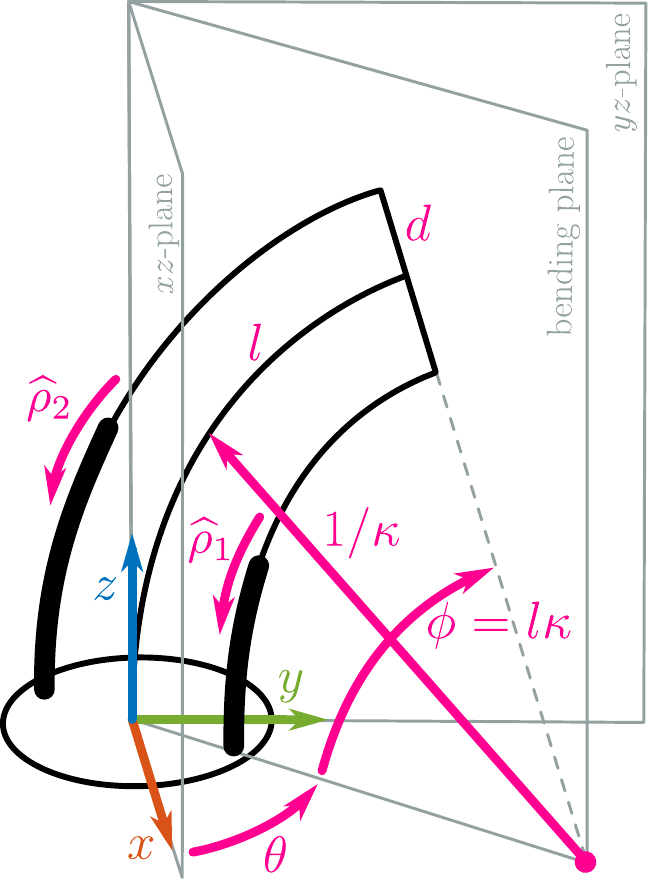}
    \vspace*{-0.75em}
    \caption{
        Surrogate displacement-actuated continuum robot.
        % It is a planner 
        }
    \label{fig:surrogate_dacr}
    \vspace*{-2em}
\end{figure}

We consider a displacement-actuated continuum robot with differential actuation and an additional rotational base, see Fig.~\ref{fig:surrogate_dacr}.
Due to the rotation $\theta$, the joint representation \eqref{eq:rho} does not apply to this toy example.
Instead, we denote the value of each displacement-actuated joint as $\widehat{\rho}_1$ and $\widehat{\rho}_2$, respectively.

As mentioned by Grassmann \textit{et al.}, the maximum and minimum values are aligned with the bending plane, assuming infinite large $n$, \textit{i.e.,} infinite many joints around the center-line.
In our case, $\widehat{\rho}_1$ is associated with the maximum value, whereas $\widehat{\rho}_2$ is associated with the minimum value.
Due to \eqref{eq:sum_rho}, $\widehat{\rho}_2 = -\widehat{\rho}_1$.
Furthermore, assuming that $\widehat{\rho}_1$ and $\widehat{\rho}_2$ relates to the Clarke coordinates, we can immediately conclude $\widehat{\rho}_1 = |\rhoclarke|$ and $\widehat{\rho}_2 = -|\rhoclarke|$.
This observation aligns with the concept of virtual displacements \cite{FirdausVadali_AIR_2023, GrassmannSenykBurgner-Kahrs_arXiv_2024}.

\begin{figure}
    \centering
    \vspace*{0.75em}
    \includegraphics[width=0.8\columnwidth]{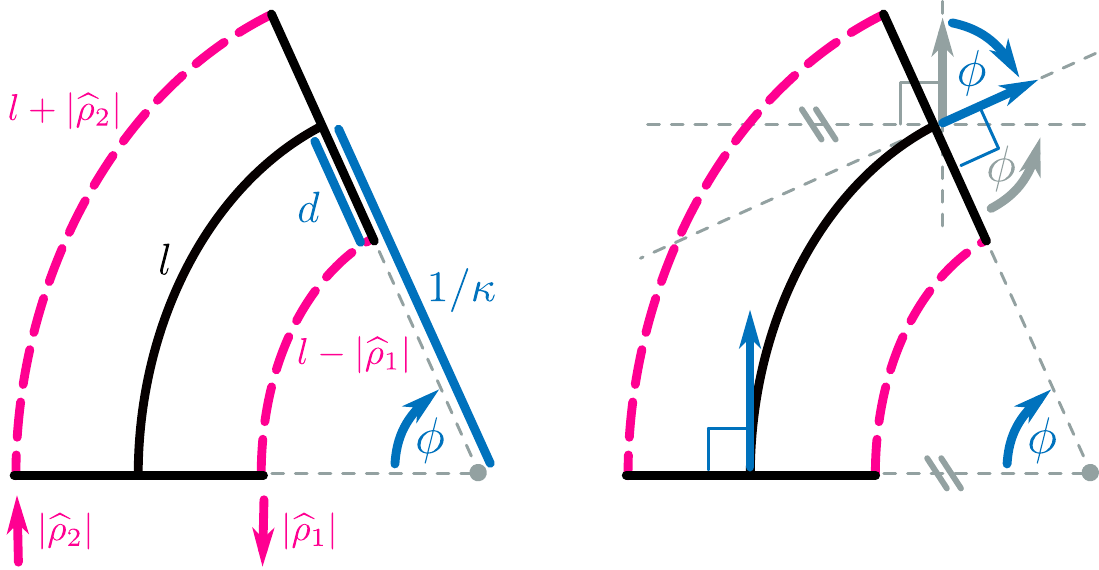}
    \vspace*{-1em}
    \caption{
        Arc length of the surrogate displacement-actuated continuum robot.
        For convenience, the absolute values of $\widehat{\rho}_1$ and $\widehat{\rho}_2$ are used.
        The bending angle $\phi$ is identical to the tip orientation.
        Note that the bending plane angle $\theta$ is other tip orientation.
        Both angles are well-known and reflected in the rotation matrix of the kinematics for the constant curvature model.
        }
    \label{fig:arc_length}
    \vspace*{-1.5em}
\end{figure}

We gain more insight by looking into the planar representation utilizing the bending plane illustrated in Fig.~\ref{fig:arc_length}, \textit{cf.} Fig.~\ref{fig:surrogate_dacr}.
Assuming constant curvature, we can relate each arc length to each other, \textit{e.g.,} $l - \widehat{\rho}_1 = \left(1/\kappa - d\right)\phi$.
Solving for $\widehat{\rho}_1$ results in $\widehat{\rho}_1 = |\rhoclarke|$, where $\phi = l\kappa$ is used.
Furthermore, we can identify that bending plane angle $\phi$ relates to the tip orientation, which is a known observation \cite{SimaanTaylorFlint_ICRA_2004}.
Note that constant curvature implicit assumes that the line associated with $\widehat{\rho}_1$ and $\widehat{\rho}_2$, respectively, is a fully constrained path, where, loosely speaking, the distance $d$ to the backbone is constant.
In fact, each line is a parallel curve, also known as an offset curve.
A parallel curve depends only on the local curvature and its distance to the curve \cite{Pham_CAD_1992}.
In Fig.~\ref{fig:arc_length}, two parallel curves with distance $d$ are depicted.

\begin{figure}
\centering
    \vspace*{0.75em}
    \includegraphics[width=0.8\columnwidth]{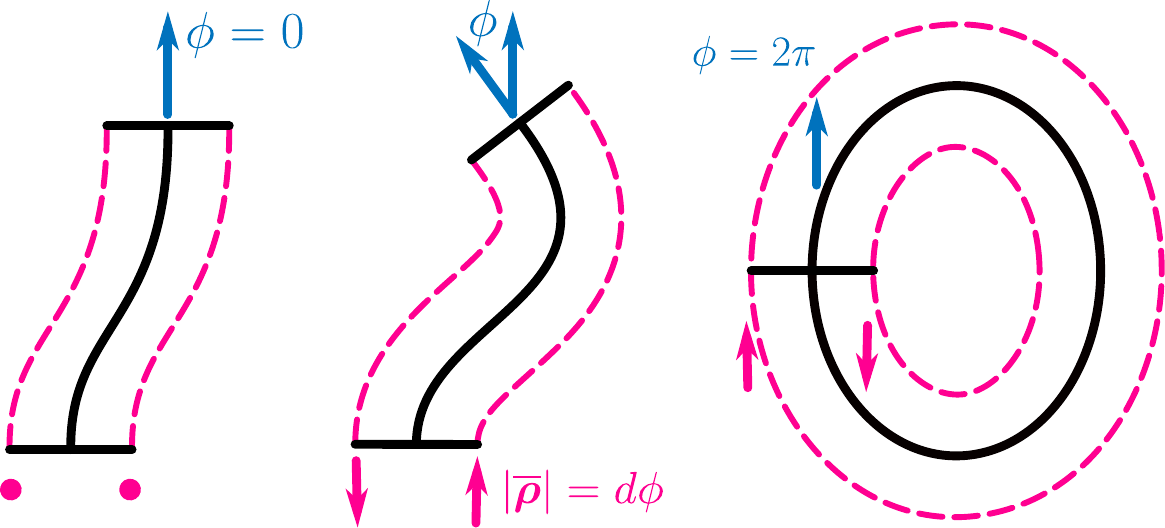}
    \vspace*{-1em}
    \caption{
    Parallel curves.
    Dashed lines are parallel curves and the solid lines between them are the center-line.
    The displacement depends on the angle $\phi$ and the distance $d$.
    (left) \SI{2}{dof} case of Cavalieri's principle, where the tip and base orientations are identical, resulting in $\widehat{\rho}_i = 0$.
    (middle) parallel curves resulting in $\widehat{\rho}_i = \pm d\phi$.
    (right) full circle, where the tip and base poses are identical, resulting in $\widehat{\rho}_i = \pm 2\pi d$.
    }
    \label{fig:parallel_curve}
    \vspace*{-1.5em}
\end{figure}

Using the formula for the arc length of a parallel curve stated by \textit{Loria} \cite{LoriaSchuette_book_1902}, a property can be derived stating that the arc length of the curve is equal to the mean of the arc lengths of both parallel curves.
In our case, using the quantities in Fig.~\ref{fig:arc_length} with sign convention depicted in Fig.~\ref{fig:surrogate_dacr} leads to 
\begin{align}
    l 
    = 
    \left(\left(l - \widehat{\rho}_1\right) + \left(l - \widehat{\rho}_2\right)\right)/2
    = 
    l - \left(\widehat{\rho}_1 + \widehat{\rho}_2\right)/2
    .
    \nonumber
\end{align}
Further algebraic manipulation results in $\widehat{\rho}_1 + \widehat{\rho}_2 = 0$ which is precisely the displacement constraint \eqref{eq:sum_rho} for $n = 2$.
The derivation using parallel curves shows that \eqref{eq:sum_rho} is not a consequence of the constant curvature assumption.
Figure~\ref{fig:parallel_curve} provides additional examples for visual aid.
Moreover, since the displacement $|\rhoclarke|$ depends on the angle $\phi$ (obtained by integrating the local curvature along the curve) and the distance $d$, \textit{cf.}, \cite{LoriaSchuette_book_1902} and \cite{SimaanTaylorFlint_ICRA_2004}, we can write
\begin{align}
    |\rhoclarke|
    &=
    d\phi
    \label{eq:rho_clarke_modulus_alt}
\end{align}
instead of \eqref{eq:rho_clarke_modulus}, where $\phi$ is the tip orientation and applicable beyond commonly used constant curvature assumption.

Moreover, the modulus of $\rhoclarke$ can be defined independently of the constant curvature assumption \cite{WebsterJones_IJRR_2010} and $d$.
We can find 
\begin{align}
    |\rhoclarke|
    &=
    \sqrt{{2}/{n}}\sqrt{\rhovec\transpose\rhovec}
    \label{eq:rho_clarke_modulus_rho}
\end{align}
by equating the dot products of \eqref{eq:rho_clarke} and \eqref{eq:inverse}.
Furthermore, the property $\MP\transpose = (2/n)\MPinv$ is used, which can be derived by inspection of \eqref{eq:MP} and \eqref{eq:MP_inverse}, \textit{cf.} \cite{GrassmannBurgner-Kahrs_ICRA_EA_2024}.

\subsection{Encoder-Decoder Architecture}
\label{sec:encoder-decoder}

\begin{figure}
    \centering
    \vspace*{0.75em}
    \includegraphics[width=\columnwidth]{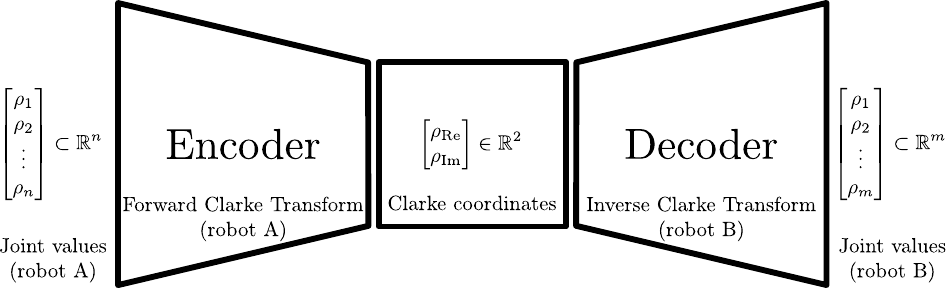}
    \vspace*{-1.5em}
    \caption{
        Encoder-decoder architecture.
        Joint values of one robot type (robot A) can be transformed into joint values of a different robot type (robot B).
        The latent space representation is encoded as Clarke coordinates.
        Note that the compression is lossless allowing joint values to be uniquely reconstructed from the Clarke coordinates. 
        (Image credit: Grassmann \& Burgner-Kahrs \cite{GrassmannBurgner-Kahrs_arXiv_2024})
        }
    \label{fig:encoder-decoder}
    \vspace*{-1.5em}
\end{figure}

The encoder-decoder architecture \cite{GrassmannBurgner-Kahrs_arXiv_2024} utilizes the fact that the Clarke coordinates $\rhoclarke$ can be seen as latent space representation.
In short, we can use \eqref{eq:forward} to map $\rhovec_{\left(\text{robot A}\right)}$ of \textit{robot A} to the same $\rhoclarke$ that are obtained using \eqref{eq:inverse} from $\rhovec_{\left(\text{robot B}\right)}$ of \textit{robot B}.
This leads to 
\begin{align}
    \rhovec_{\left(\text{robot B}\right)} = 
    \underbrace{
    {\MPinv}_{\left(\text{robot B}\right)}
    }_{\substack{\text{decoder}}}
    \underbrace{
    {\MP}_{\left(\text{robot A}\right)}
    }_{\substack{\text{encoder}}}
    \rhovec_{\left(\text{robot A}\right)},
    \label{eq:encoder-decoder}
\end{align}
which is called \textit{encoder-decoder architecture} \cite{GrassmannBurgner-Kahrs_arXiv_2024}.
Figure~\ref{fig:encoder-decoder} visualizes this approach.
A more general approach is presented in \cite{GrassmannBurgner-Kahrs_arXiv_2024} as well.

\section{Feasible Joint Values}
\label{sec:sampling}

We directly sample Clarke coordinates $\rhoclarke_\mathcal{U}$ using a rejection-free sampling method \cite{GrassmannSenykBurgner-Kahrs_arXiv_2024}.
Considering the argument \eqref{eq:rho_clarke_argument} and modulus \eqref{eq:rho_clarke_modulus_alt} of $|\rhoclarke|$, we only need to sample
\begin{align}
    |\rhoclarke_\mathcal{U}|
    =
    \phi_\text{max} d\, \mathcal{U}\left[0; 1\right]
    \quad\text{and}\quad
    \arg \rhoclarke_\mathcal{U}
    =
    \theta_\text{max}\, \mathcal{U}\left[-1; 1\right)
    ,
    \nonumber
\end{align}
where $\phi_\text{max}$ and $\theta_\text{max}$ define the maximum value for the tip orientations, see Fig.~\ref{fig:surrogate_dacr} and Fig.~\ref{fig:arc_length}.
Each value is drawn from a uniform distribution $\mathcal{U}$.
To match the application at hand, other distributions and ranges might be considered.
Furthermore, if more samples are required, this sampling method can be vectorized.

Setting $\phi_\text{max}$ is more convenient than defining a maximum curvature $\kappa_\text{max}$. 
The convenience lies in relation to the tip orientation, recap Fig.~\ref{fig:arc_length}, and the visualization using circles, where an angle describes a circle more intuitively than the corresponding curvature.
Recap the relation to virtual displacement in Sec.~\ref{sec:virtual_displacement} and Fig.~\ref{fig:parallel_curve}, it also circumvents the relation to constant curvature, \textit{i.e.}, $\phi_\text{max} = l\kappa_\text{max}$.
\section{Trajectory Generation}

In this paper, we focus on simple polynomial trajectories to lay out the basics of trajectory generation using Clarke coordinates.
First, we establish a simple point-to-point trajectory for the displacement.
Afterward, a point-to-point trajectory generator is proposed for Clarke coordinates.
Finally, the connection between both trajectory generators is pointed out.

\subsection{Trajectory Generation in Joint Space}

Consider a $\mathcal{C}^3$ smooth path primitive $s\!\left(\tau\right) \in \left[0, 1\right]$ and its derivative \textit{w.r.t.} the time primitive $\tau \in \left[0, 1\right]$ given by
\begin{align}
    s\!\left(\tau\right) =&  -20\tau^7 + 70\tau^6 - 84\tau^5 + 35\tau^4,\nonumber\\
    s'\!\left(\tau\right) =&  -140\tau^6 + 420\tau^5 - 420\tau^4 + 140\tau^3\quad\text{, and}\nonumber\\
    s''\!\left(\tau\right) =&  -840\tau^5 + 2100\tau^4 - 1680\tau^3 + 420\tau^2,\nonumber
\end{align}
respectively \cite{BiagiottiMelchiorri_Book_2008}.
The duration $T$ of the trajectory is included in the definition of the time primitive $\tau = t/T$.
Using the boundaries $s\!\left(1\right) = 1$ and $s\!\left(0\right) = s'\!\left(0\right) = s'\!\left(1\right) = s''\!\left(0\right) = s''\!\left(0\right) = 0$, the path primitive and its above derivative can be clipped to define trajectories for all $t$.
For a given $\rhovec_\text{start} \in \mathbf{Q}$ and $\rhovec_\text{goal} \in \mathbf{Q}$, we can now define a simple trajectory as
\begin{align}
    \rhovec_\mathrm{d}\!\left(t\right) = \rhovec_\text{start} + \left( \rhovec_\text{goal} - \rhovec_\text{start}\right)s\!\left(t\right).
    \label{eq:trajectory}
\end{align}
It is important to show that, at every time step, a valid displacement is generated.
For this, checking \eqref{eq:sum_rho} results in
\begin{align}
    \sum_{i=1}^{n}\rho_{i, \mathrm{d}}\!\left(t\right) 
    &= \sum_{i=1}^{n}\left(\rho_{i, \text{start}} + \left( \rho_{i, \text{goal}} - \rho_{i, \text{start}}\right)s\!\left(t\right)\right)
    \nonumber\\
    &= \sum_{i=1}^{n}\rho_{i, \text{start}} + s\!\left(t\right)\sum_{i=1}^{n}\rho_{i, \text{goal}} - s\!\left(t\right)\sum_{i=1}^{n}\rho_{i, \text{start}}
    \nonumber\\
    &= 0,
    \nonumber
\end{align}
because $\rhovec_\text{start} \in \mathbf{Q}$ and $\rhovec_\text{goal} \in \mathbf{Q}$ obey \eqref{eq:sum_rho}.
This simple check shows that $\rhovec_\mathrm{d}\!\left(t\right) \in \mathbf{Q}$ for all $t$.

For convenience, we define the maximal absolute value of all entries in $\rhovec_\text{goal} - \rhovec_\text{start}$ as
\begin{align}
    \Delta\rho = \| \rhovec_\text{goal} - \rhovec_\text{start} \|_\infty,
    \label{eq:rho_infty}
\end{align}
which is a common choice for trajectory generation of positions in Task space.
The quantity $\Delta\rho$ is important to synchronize all $n$ trajectories of \eqref{eq:trajectory}.

To consider the kinematic constraints given by $\vmax$ for the maximal velocity and $\amax$ for the maximal acceleration, each time at the maximal value of $s'\!\left(t\right)$ and $s''\!\left(t\right)$ is determined.
By choosing a large enough duration $T$, the kinematic constraints can be considered.
Therefore, the duration $T$ is
\begin{align}
    T = \max \left\{\dfrac{35\Delta\rho}{16\vmax},\, \sqrt{\dfrac{84\Delta\rho}{5\sqrt{5}\amax}},\, T_\text{user}\right\},
    \label{eq:duration}
\end{align}
where $T_\text{user}$ is defined by a user.
For $T_\text{user} = 0$, the executed trajectory \eqref{eq:trajectory} always reaches one of the kinematic constraints.
Furthermore, due to the consideration of \eqref{eq:rho_infty} in \eqref{eq:duration}, all trajectories in \eqref{eq:trajectory} are synchronized, recall $\tau = t/T$.

\subsection{Trajectory Generation using Clarke Coordinates}
\label{sec:trajectory_generation_Clarke}

Until now, all trajectories generated with geometric scaling using path primitives $s\!\left(t\right)$ have shown to be feasible.
Mapping the trajectory \eqref{eq:trajectory} onto the manifold leads to
\begin{align}
    \rhoclarke_\mathrm{d}\!\left(t\right)  &= \MP\rhovec_\mathrm{d}\!\left(t\right)\nonumber\\
    &= \MP\left(\rhovec_\text{start} + \left( \rhovec_\text{goal} - \rhovec_\text{start}\right)s\!\left(t\right)\right)\nonumber\\
    &= \MP\rhovec_\text{start} + s\!\left(t\right)\MP\rhovec_\text{goal} - s\!\left(t\right)\MP\rhovec_\text{start}\nonumber\\
    &= \rhoclarke_\text{start} + s\!\left(t\right)\rhoclarke_\text{goal} - s\!\left(t\right)\rhoclarke_\text{start},\nonumber
\end{align}
where the last step results in the formulation
\begin{align}
    \rhoclarke_\mathrm{d}\!\left(t\right)  =& \rhoclarke_\text{start} + \left( \rhoclarke_\text{goal} - \rhoclarke_\text{start}\right)s\!\left(t\right)
    \label{eq:trajectory_clarke}
\end{align}
for a trajectory defined on the manifold.
Note that \eqref{eq:trajectory} is a straight line in the joint space $\mathbf{Q}$ and \eqref{eq:trajectory_clarke} is also a straight line on the manifold.
The derivation of \eqref{eq:trajectory_clarke} shows that it is consistent with a trajectory \eqref{eq:trajectory} defined in $\mathbf{Q}$.

To consider the kinematic constraints defined on the manifold, we can adapt the time $T$ by substituting 
\begin{align}
    \Delta\rho &\longleftarrow \Delta\bar{\rho}, \nonumber\\
    \vmax\rho &\longleftarrow \bar{v}_\text{max}\quad\text{, and} \nonumber\\
    \amax &\longleftarrow \bar{a}_\text{max} \nonumber
\end{align}
into \eqref{eq:duration}, where the displacement $\Delta\bar{\rho}$ is defined by 
\begin{align}
    \Delta\bar{\rho} = \| \rhoclarke_\text{goal} - \rhoclarke_\text{start} \|_\infty.
    \label{eq:rhoclarke_Linf}
\end{align}
While the definition of the kinematic constraints defined in the joint space $\mathbf{Q}$ are meaningful due to the physical relation to the displacement-actuated continuum robot, the equivalent kinematic constraints used for \eqref{eq:trajectory_clarke}, \textit{i.e.}, $\bar{v}_\text{max}$ and $\bar{a}_\text{max}$ are obscure.
Therefore, there is a need to establish the relation between those two sets of kinematic constraints.

\subsection{Relation between the Trajectory Generators}

The physical meaning of the kinematic constraints defined in $\mathbf{Q}$ depends on the used displacement-actuated continuum robot.
For example, a tendon-driven continuum robot uses tendons that actuated with a winch or a threaded drum attached to a motor.
Another exemplary displacement-actuated continuum robot is one with multiple secondary backbones actuated with linear actuators. 
In those cases, the physical meaning of $\vmax$ and $\amax$ are straightforward.
For a soft robot actuated with bellows, the relationship to $\vmax$ and $\amax$ is slightly more complicated.

To relate both sets of kinematic constraints, we find the extrema of time derivatives of \eqref{eq:trajectory} and \eqref{eq:trajectory_clarke}, respectively.
Since the trajectory is synchronized, it is sufficient to look at the $i\textsuperscript{th}$ entry of \eqref{eq:trajectory} and the real part of \eqref{eq:trajectory_clarke}.
We can find
\begin{align}
    \max\dfrac{\mathrm{d}}{\mathrm{d}t}\rho_{i, \mathrm{d}}\!\left(t\right) &= \dfrac{\Delta\rho}{T}\dfrac{35}{16} = \vmax \quad\text{and}
    \nonumber
    \\
    \max\dfrac{\mathrm{d}^2}{\mathrm{d}^2t}\rho_{i, \mathrm{d}}\!\left(t\right) &= \dfrac{\Delta\rho}{T^2}\dfrac{84}{5\sqrt{5}} = \amax
    \nonumber
\end{align}
for \eqref{eq:trajectory}, whereas the extrema for \eqref{eq:trajectory_clarke} are
\begin{align}
    \max\dfrac{\mathrm{d}}{\mathrm{d}t}\rho_{\text{Re}, \mathrm{d}}\!\left(t\right) &= \dfrac{\Delta\bar{\rho}}{T}\dfrac{35}{16} = \bar{v}_\text{max} \quad\text{and}
    \nonumber
    \\
    \max\dfrac{\mathrm{d}^2}{\mathrm{d}^2t}\rho_{\text{Re}, \mathrm{d}}\!\left(t\right) &= \dfrac{\Delta\bar{\rho}}{T^2}\dfrac{84}{5\sqrt{5}} = \bar{a}_\text{max}
    .
    \nonumber
\end{align}
Afterward, $T$ is equated and the resulting ratios are rearranged.
The relation between the kinematic constraint is
\begin{align}
    \bar{v}_\text{max} = \vmax\dfrac{\Delta\bar{\rho}}{\Delta\rho} 
    \quad\text{and}\quad
    \bar{a}_\text{max} = \amax\dfrac{\Delta\bar{\rho}}{\Delta\rho}
    ,
    \nonumber
\end{align}
respectively.
However, the above relations rely on the evaluation of \eqref{eq:rho_infty} and \eqref{eq:rhoclarke_Linf}.
To overcome this, we can exploit 
\begin{align}
    \| \boldsymbol{x} \|_\infty \leq \| \boldsymbol{x} \|_2 \leq \sqrt{n}\| \boldsymbol{x} \|_\infty 
    \nonumber
\end{align}
for any vector $\boldsymbol{x}$.
Now, we can utilize \eqref{eq:rho_clarke_modulus_rho} and derive lower and upper bounds.
The lower bounds are given by
\begin{align}
    \bar{v}_\text{max} 
    &\geq
    \sqrt{\dfrac{2}{n}} \, \dfrac{\| \bar{\rhovec}_\text{goal} - \bar{\rhovec}_\text{start} \|_\infty}{\| \bar{\rhovec}_\text{goal} - \bar{\rhovec}_\text{start} \|_2}\vmax
    \geq
    \dfrac{\sqrt{2}}{n} \vmax
    \quad\text{and}
    \nonumber
    \\
    \bar{a}_\text{max} 
    &\geq
    \sqrt{\dfrac{2}{n}} \, \dfrac{\| \bar{\rhovec}_\text{goal} - \bar{\rhovec}_\text{start} \|_\infty}{\| \bar{\rhovec}_\text{goal} - \bar{\rhovec}_\text{start} \|_2}\amax
    \geq
    \dfrac{\sqrt{2}}{n} \amax
    .
    \nonumber
\end{align}
Since lower values for $\bar{v}_\text{max}$ and $\bar{a}_\text{max}$ will satisfy the kinematic constraints in joint space, we set
\begin{align}
    \bar{v}_\text{max} &\leftarrow \sqrt{\dfrac{2}{n}} \, \dfrac{\| \bar{\rhovec}_\text{goal} - \bar{\rhovec}_\text{start} \|_\infty}{\| \bar{\rhovec}_\text{goal} - \bar{\rhovec}_\text{start} \|_2}\vmax
    \quad\text{and}\quad
    \label{eq:kinematic_constraints_relation_vmax}
    \\
    \bar{a}_\text{max} &\leftarrow \sqrt{\dfrac{2}{n}} \, \dfrac{\| \bar{\rhovec}_\text{goal} - \bar{\rhovec}_\text{start} \|_\infty}{\| \bar{\rhovec}_\text{goal} - \bar{\rhovec}_\text{start} \|_2}\amax
    ,
    \label{eq:kinematic_constraints_relation_amax}
\end{align}
allowing to define the kinematic constraints for \eqref{eq:trajectory_clarke}.
\section{Control}

The Clarke transform enables the synthesis of linear controllers that utilize the $\SI{2}{dof}$ manifold.
A potential controller scheme could be a linear controller with two degrees of freedom.
Grassmann \textit{et al.} proposed two independent proportional feedback controllers with pre-compensation based on work by Morin \& Samon \cite{MorinSamson_HOR_2008}.
In Grassmann \& Burgner-Kahrs \cite{GrassmannBurgner-Kahrs_arXiv_2024}, two independent PD controllers are utilized, where the controller gains are defined as $\boldsymbol{K}_\mathrm{P} = k_\mathrm{P}\Imat_{2 \times 2}$ and $\boldsymbol{K}_\mathrm{D}  = k_\mathrm{D}\Imat_{2 \times 2}$.
Non-diagonal $\boldsymbol{K}_\mathrm{P} \in \mathbb{R}^{2 \times 2}$ and $\boldsymbol{K}_\mathrm{D} \in \mathbb{R}^{2 \times 2}$ are also possible.

\begin{figure}
    \centering
    \vspace*{0.75em}
    \includegraphics[width=\columnwidth]{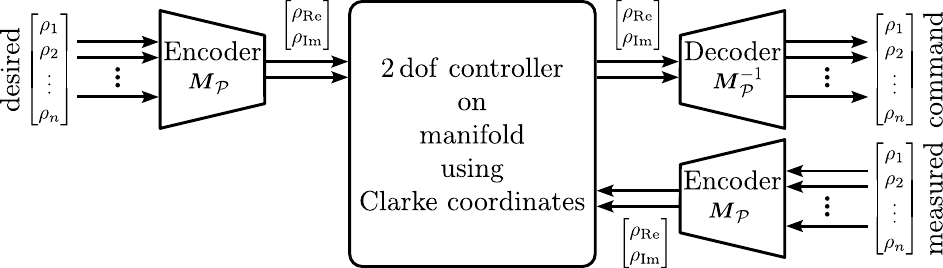}
    \vspace*{-1.75em}
    \caption{
        Controller scheme utilizing Clarke transform.
        For the sake of compactness, we omit potential derivative terms such as velocity.
        }
    \label{fig:controller_encoder-decoder}
\end{figure}

Figure~\ref{fig:controller_encoder-decoder} illustrates a general controller scheme using parts of the encoder-decoder architecture \eqref{eq:encoder-decoder}.
In this general framework, controllers beyond the PID controller can be used including non-linear controller schemes.
In a framework solely based on Clarke coordinates, the encoder on the left side depicted in Fig.~\ref{fig:controller_encoder-decoder} can be omitted, where the desired Clarke coordinates are provided by a trajectory generator.
However, it can be used as an interface for external frameworks outputting $\rhovec$ and its time derivatives.

\section{Framework and Demonstration in Simulation}

\begin{figure*}[thb]
    \centering
    \vspace*{0.75em}
    \includegraphics[width=\textwidth]{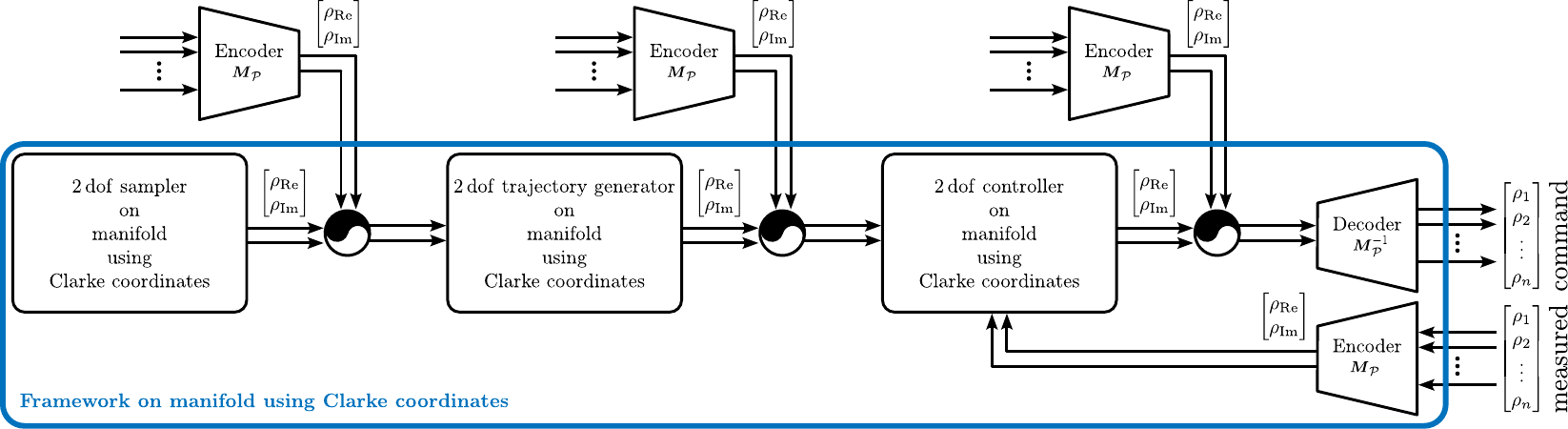}
    \vspace*{-2em}
    \caption{
        Framework based on Clarke coordinates. 
        Each of the components of the framework, \textit{i.e.}, sampling, trajectory generation, and control, are solely defined in terms of the Clarke coordinates.
        Using an encoder as an interface, existing frameworks can be included and reused.
        A switch depicted as a circular symbol enables or disables external inputs.
        %Further decoders to redirect the sampler or trajectory generator output are omitted.
        The right-hand side represents the interface with a displacement-actuated continuum robot.
        }
    \label{fig:framework}
    \vspace*{-1.5em}
\end{figure*}

Now that the planner, trajectory generator, and controller have been defined by using Clarke coordinates, the framework can be assembled.
Figure~\ref{fig:framework} illustrates the framework with additional encoders enabling the possibility to reuse components from existing frameworks.

The framework is parameterized by the boundaries of the sampler used as planner, the kinematic constraints of the trajectory generator, the controller gain of the controller scheme, as well as the kinematic design parameters for the encoder and decoder.
Table~\ref{tab:framework_parameters} lists the parameters and their values.
For the demonstration, the planner is the simple random joint generator providing feasible joint values as described in Sec.~\ref{sec:sampling}.
The used point-to-point trajectory generator is described in Sec.~\ref{sec:trajectory_generation_Clarke}.
For the controller, we use two independent P controllers with pre-compensation \cite{MorinSamson_HOR_2008} acting on the difference between the measured and desired Clarke coordinates. 
The output of the controller is passed through a decoder to a displacement-actuated continuum robot in simulation.

\begin{table}
	%\vspace*{2mm}
	\centering
	\caption{
        Parameters of the framework.
    }
	\label{tab:framework_parameters}
	\vspace*{-1em}
	\begin{tabular}{@{} r rr p{4.25cm} @{}}
		\toprule
		\multicolumn{1}{N}{Variable}
		& \multicolumn{1}{N}{Value}
		& \multicolumn{1}{N}{Unit}
		& \multicolumn{1}{N}{Description}\\
		\cmidrule(r){1-1}
		\cmidrule(lr){2-2}
		\cmidrule(lr){3-3}
		\cmidrule(l){4-4}
	    $\phi_\text{max}$ & $2\pi/3$ & \SI{}{rad} & Related to the maximum curvature $\phi_\text{max} =l\kappa_\text{max}$ \\[1.5em]
	    $\theta_\text{max}$ & $\pi$ & \SI{}{rad} & Symmetric maximum bending plane angle \\[.5em]
	    $\vmax$ & \num{0.01} & \SI{}{m/s} & Maximum velocity of $\rho_i\!\left(t\right)$\\[.5em]
	    $\amax$ & \num{0.01} & \SI{}{m/s^2} & Maximum acceleration of $\rho_i\!\left(t\right)$ \\[.5em]
	    $\bar{v}_\text{max}$ & -- & \SI{}{m/s} & Maximum velocity of $\bar{\rhovec}$ is predefined by $\vmax$ and \eqref{eq:kinematic_constraints_relation_vmax}\\[1.5em]
	    $\bar{a}_\text{max}$ & -- & \SI{}{m/s^2} & Maximum acceleration of $\bar{\rhovec}$ is predefined by $\amax$ and \eqref{eq:kinematic_constraints_relation_amax}\\[1.5em]
	    $k_\mathrm{p}$ & \num{10} & -- & Controller gain of proportional term\\[.5em]
	    $n$ & \num{5} & -- &  Number of joints\\[.5em]
	    $l$ & \num{0.07} & \SI{}{m} & Segment length\\[.5em]
	    $\psi$ & $2\pi/n$ & \SI{}{rad} & Angular distance between adjacent joint location\\[.5em]
	    $d$ & \num{0.01} & \SI{}{m} & Distant between center-line and joint location\\
		\bottomrule
	\end{tabular}
\end{table}

\begin{figure}[thb]
    \centering
    \includegraphics[width=0.8\columnwidth]{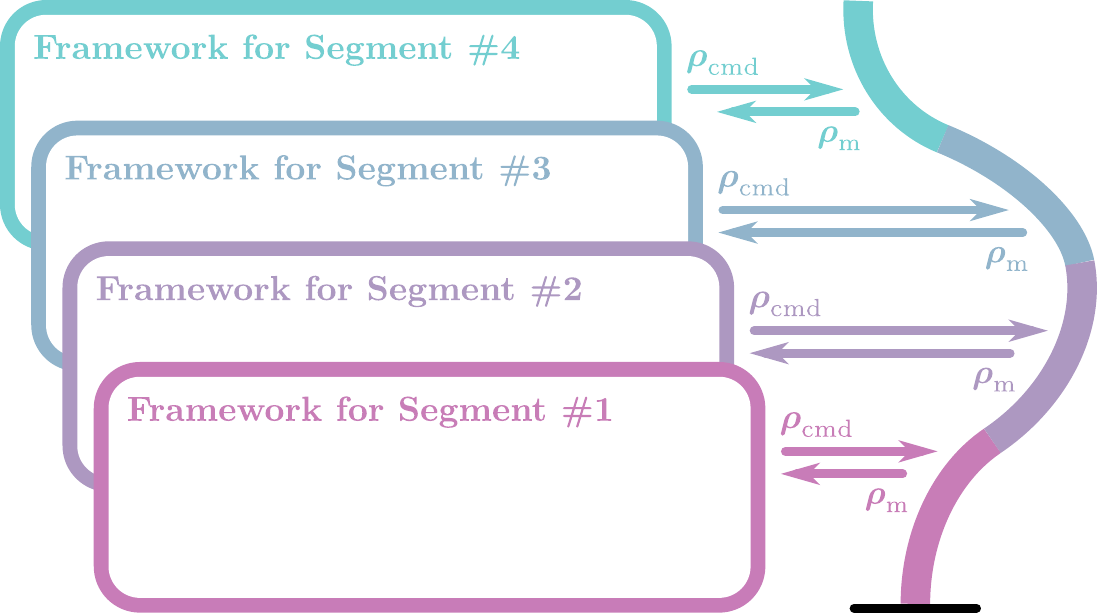}
    \vspace*{-0.75em}
    \caption{
        One segment, one framework
        }
    \label{fig:multiple_frameworks}
    \vspace*{-0.75em}
\end{figure}

For the system consisting of the framework and the coupled displacement-actuated continuum robot, the sampling time is set to \SI{10}{ms}.
To simulate the displacement-actuated continuum robot, each actuator is modeled as an interdependent first-order proportional delay element ($\text{PT}_1$) system.
The time constant for each $\text{PT}_1$ system is uniformly set to \SI{200}{ms}.
We include additive noise, drawn from a uniform distribution with an amplitude of \SI{0.1}{mm}, to measure displacement-actuated joint values $\rhovec_\mathrm{m}$.
The robot is mass-less as only the kinematic constant curvature model is considered for the visualization.
To account for several segments, each segment is treated independently and has its own instance of the framework as depicted in Fig.~\ref{fig:multiple_frameworks}, where all frameworks have the same parameters stated in Table~\ref{tab:framework_parameters}.
To synchronize all trajectories across the frameworks, the maximum duration is computed and used.

\begin{figure}
    \centering
    \includegraphics[width=0.78\columnwidth]{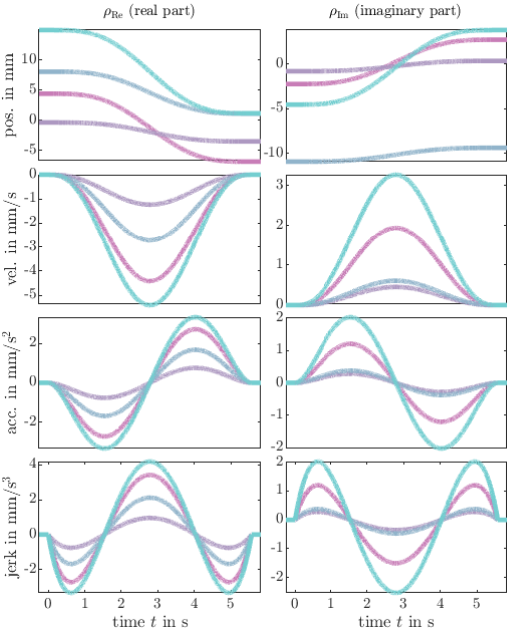}
    \vspace*{-1em}
    \caption{
        Planned trajectory using Clarke coordinates.
        }
    \label{fig:desired_clarke_coordinates}
    \vspace*{-2em}
\end{figure}

The parallelized framework illustrated in Fig.~\ref{fig:multiple_frameworks} provides displacements, which are sent to the displacement-actuated continuum robot.
Figure~\ref{fig:desired_clarke_coordinates} shows the Clarke coordinates and their time derivatives generated by the trajectory generator \eqref{eq:trajectory_clarke}.

\begin{figure*}
    \centering
    \vspace*{0.75em}
    \includegraphics[width=0.9\textwidth]{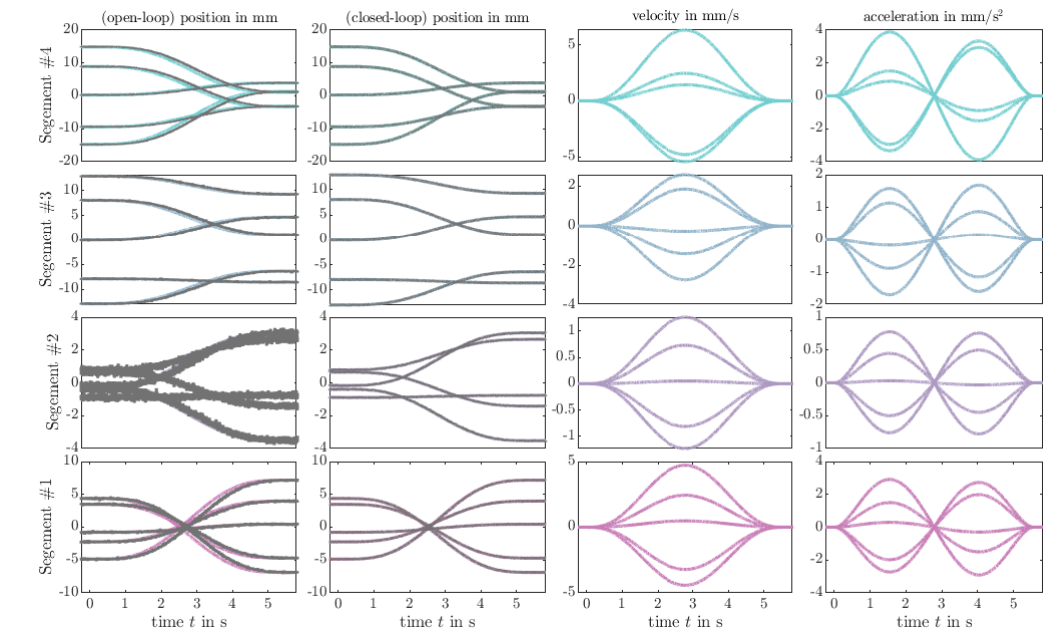}
    \vspace*{-1.25em}
    \caption{
        Displacement-control.
        ($1\textsuperscript{st}$ column) Desired path \textit{vs.} open-loop behavior of the $\text{PT}_1$ with noisy measurements.
        The noise is more pronounced for shorter displacements amplifying the noise-to-signal ratio, \textit{cf.} segment \#2 and segment \#4.
        ($2\textsuperscript{nd}$ column) Desired path \textit{vs.} closed-loop behavior.
        ($3\textsuperscript{rd}$ column) Velocity of the desired path.
        ($4\textsuperscript{th}$ column) Acceleration of the desired path that shows continuous acceleration profile, \textit{i.e.}, $\mathcal{C}^3$ smoothness.
        Measured velocity and acceleration are not used due to the high noise.
        Observe that the kinematic constraints $\vmax = \SI{10}{mm/s}$ and $\amax = \SI{10}{mm/s^2}$ are complied with.
        }
    \label{fig:results_closed_loop}
\end{figure*}

\begin{figure*}
    \centering
    \hfill
    \begin{overpic}[width=0.19\textwidth]{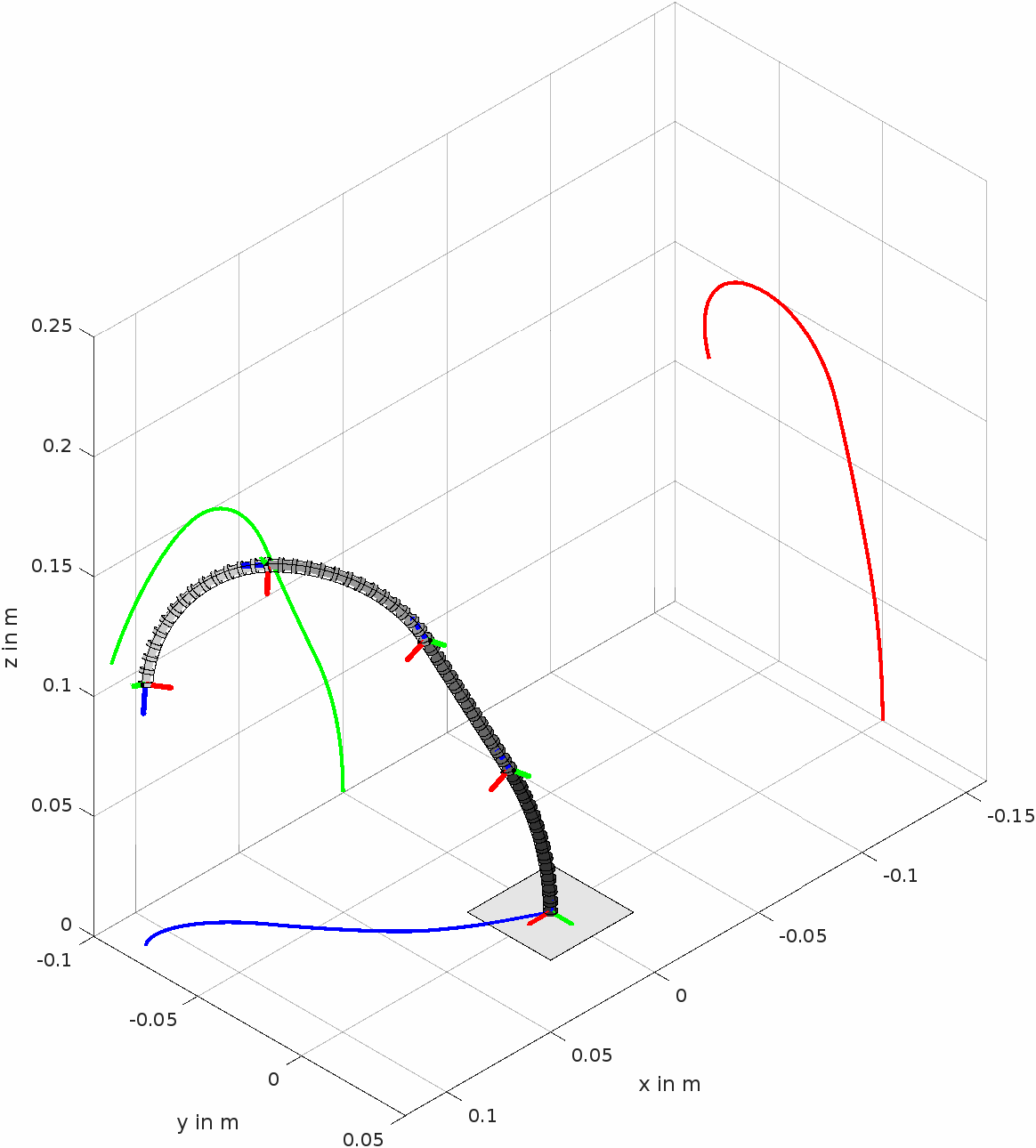}
        \put(0,85){\footnotesize$t = \SI{0.11}{s}$}
    \end{overpic}
    \begin{overpic}[width=0.19\textwidth]{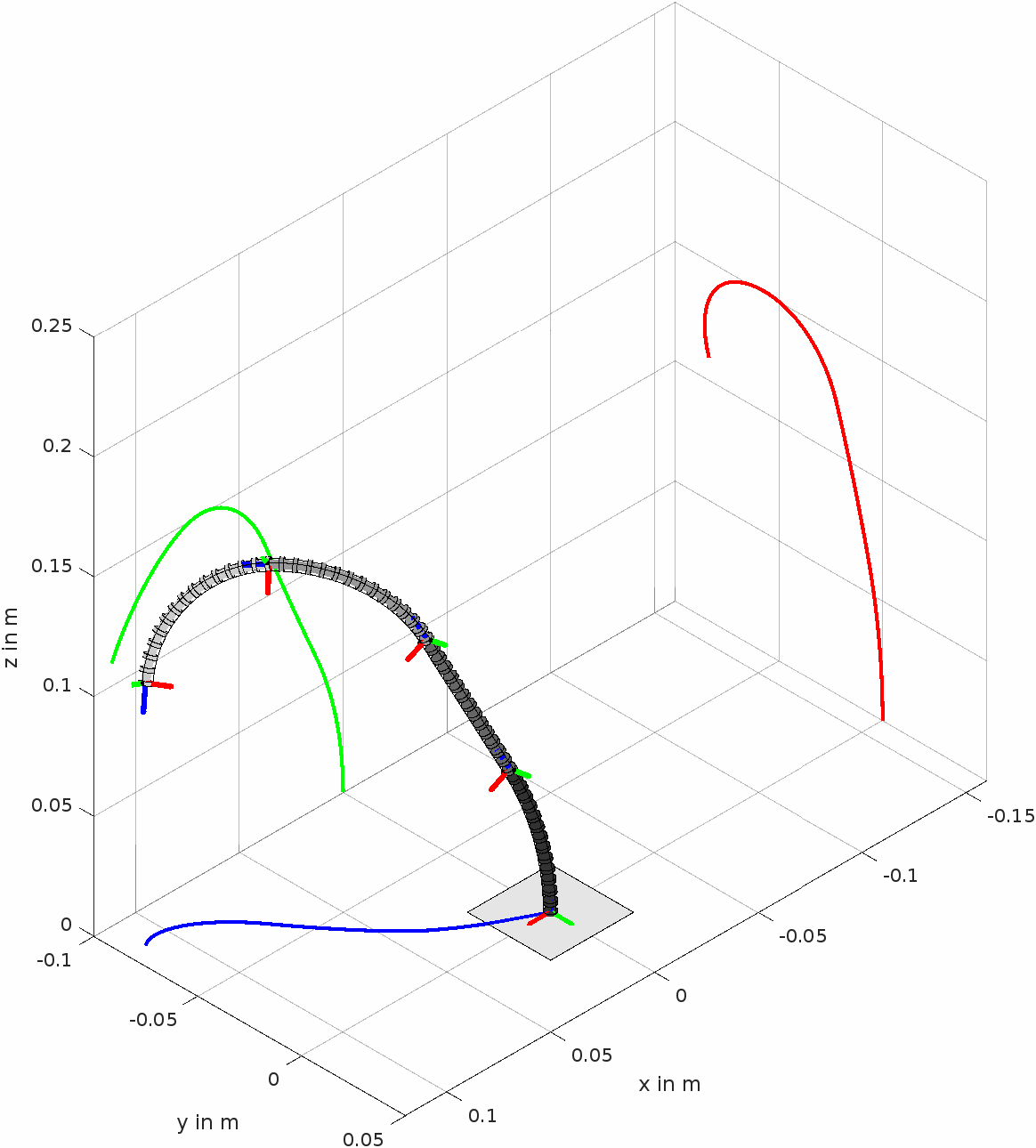}
        \put(0,85){\footnotesize$t = \SI{0.51}{s}$}
    \end{overpic}
    \begin{overpic}[width=0.19\textwidth]{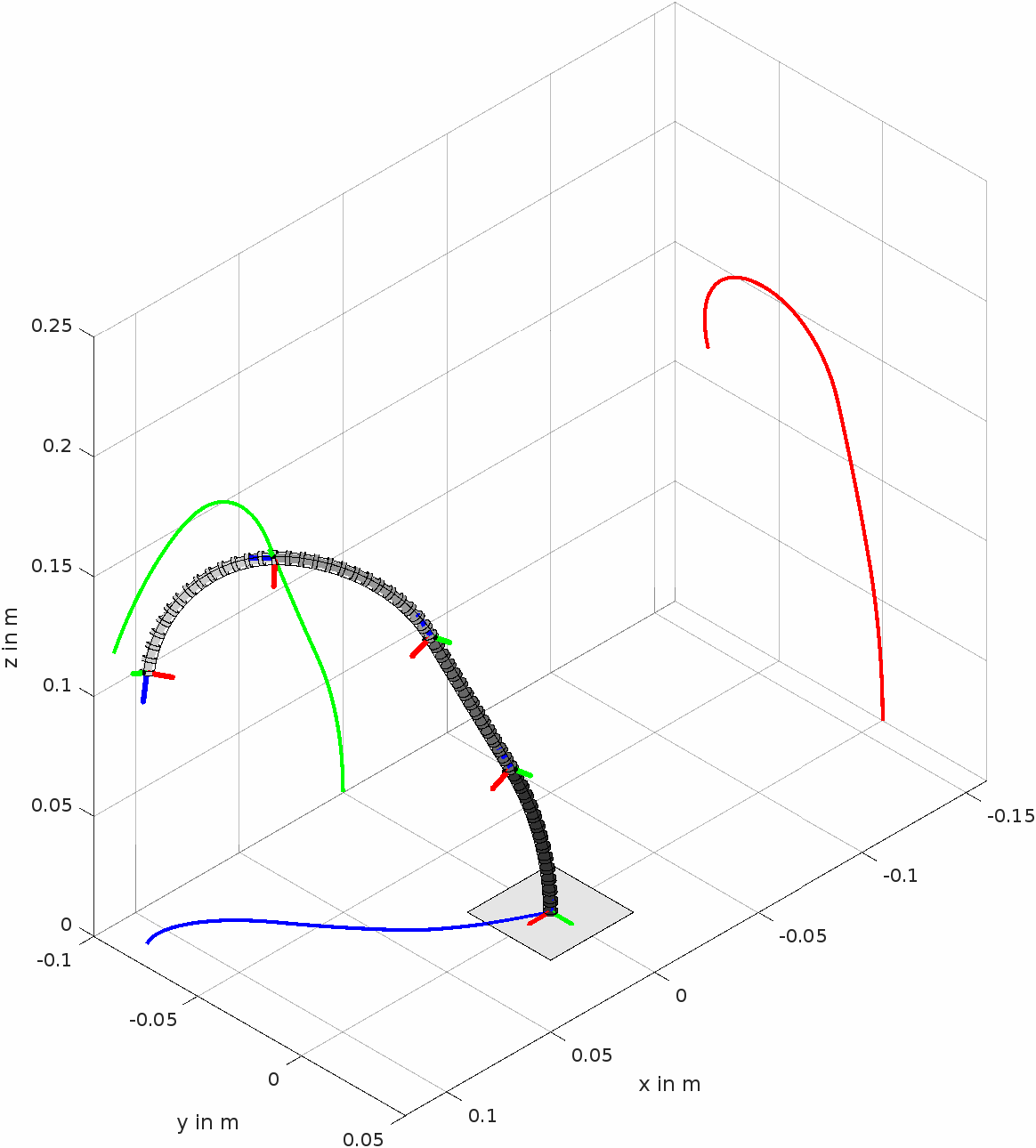}
        \put(0,85){\footnotesize$t = \SI{0.92}{s}$}
    \end{overpic}
    \begin{overpic}[width=0.19\textwidth]{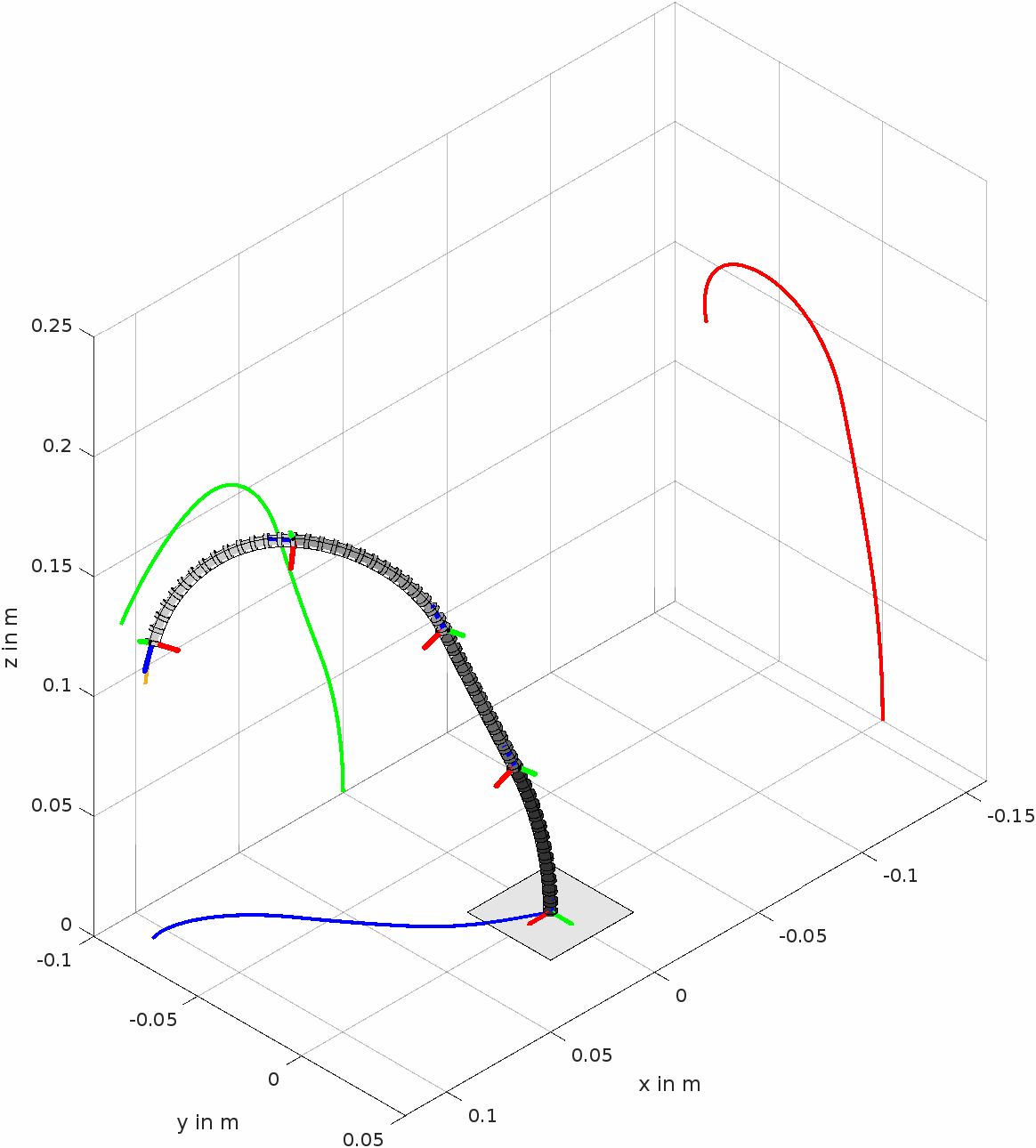}
        \put(0,85){\footnotesize$t = \SI{1.32}{s}$}
    \end{overpic}
    \begin{overpic}[width=0.19\textwidth]{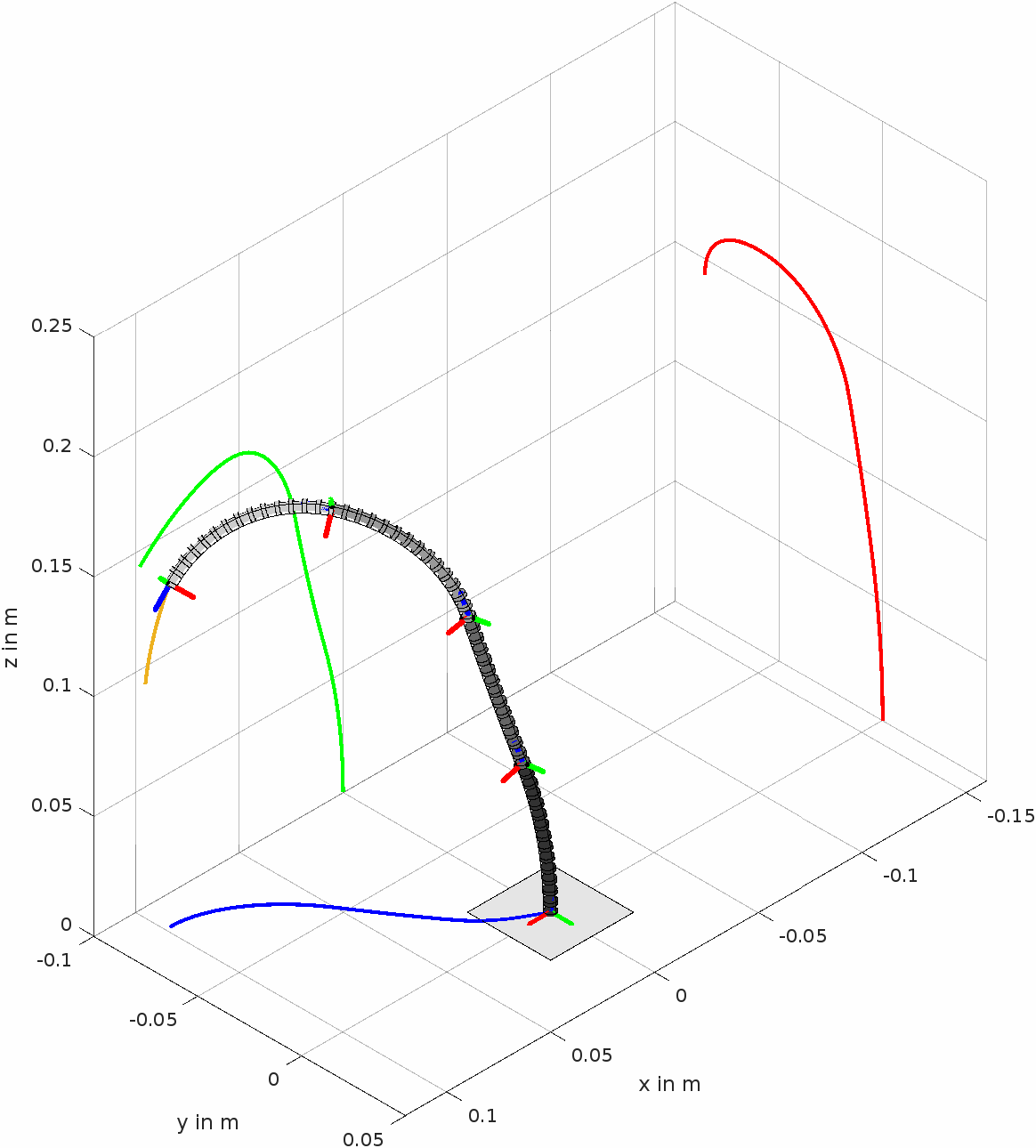}
        \put(0,85){\footnotesize$t = \SI{1.72}{s}$}
    \end{overpic}
    \hfill
    \\[0.25em]
    \hfill
    \begin{overpic}[width=0.19\textwidth]{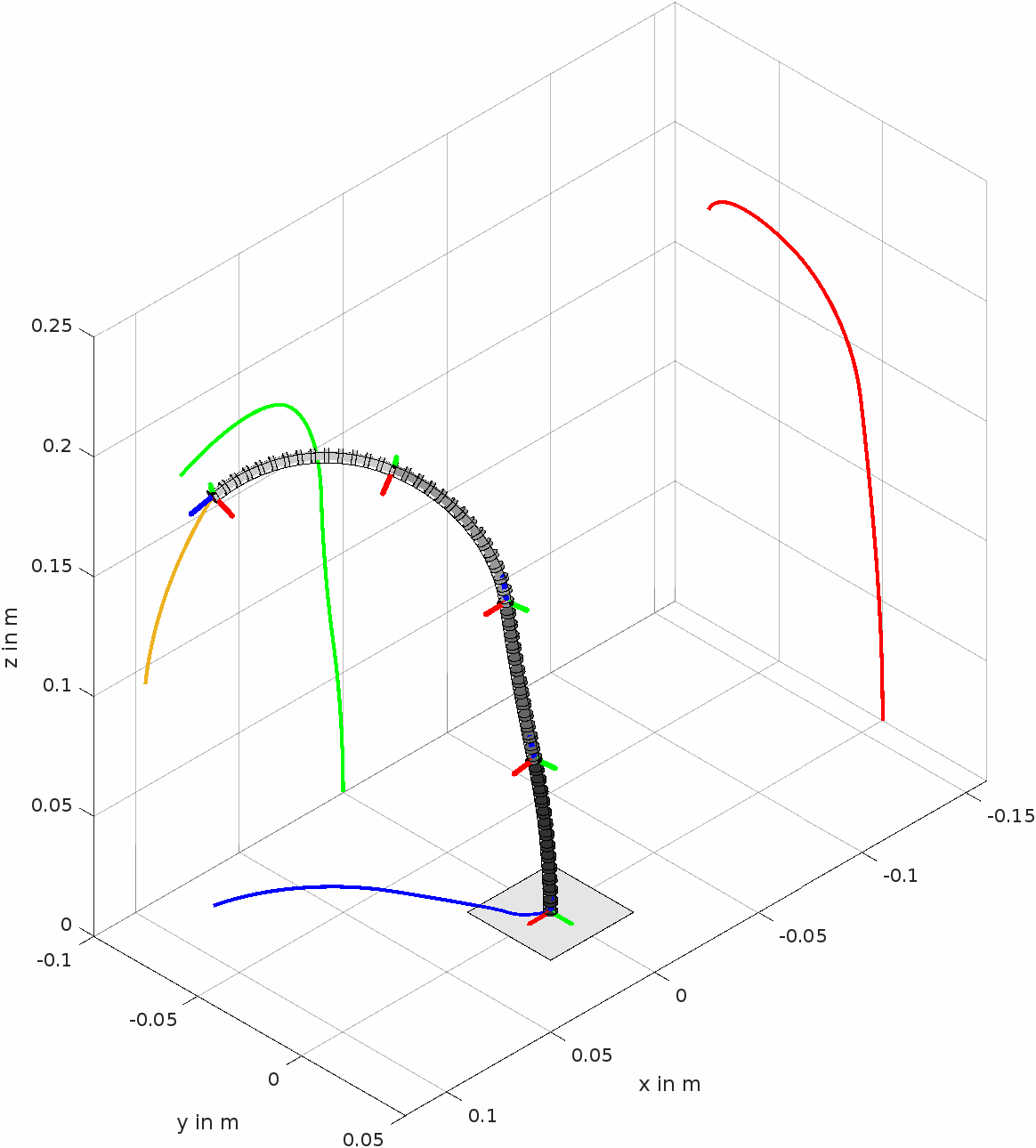}
        \put(0,85){\footnotesize$t = \SI{2.12}{s}$}
    \end{overpic}
    \begin{overpic}[width=0.19\textwidth]{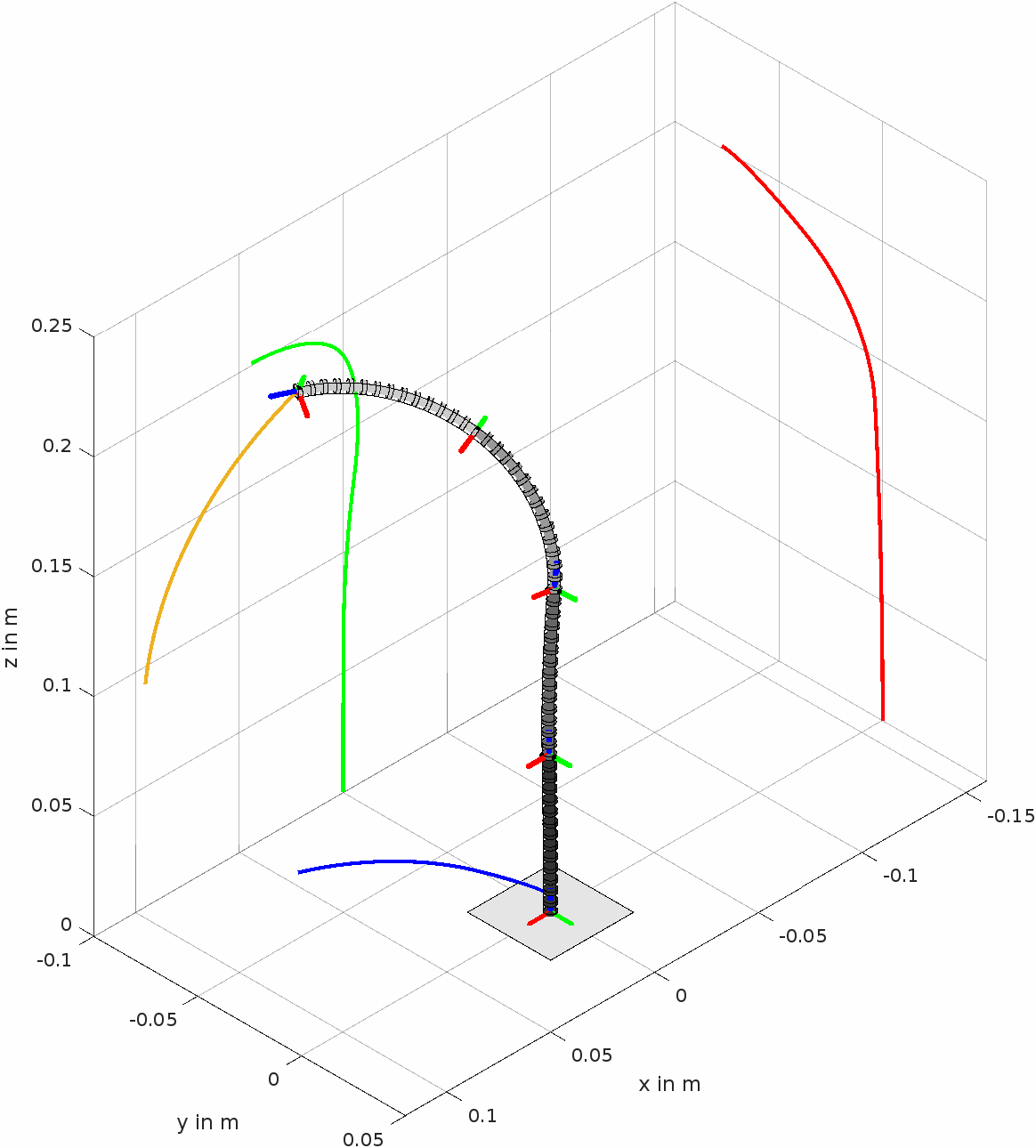}
        \put(0,85){\footnotesize$t = \SI{2.52}{s}$}
    \end{overpic}
    \begin{overpic}[width=0.19\textwidth]{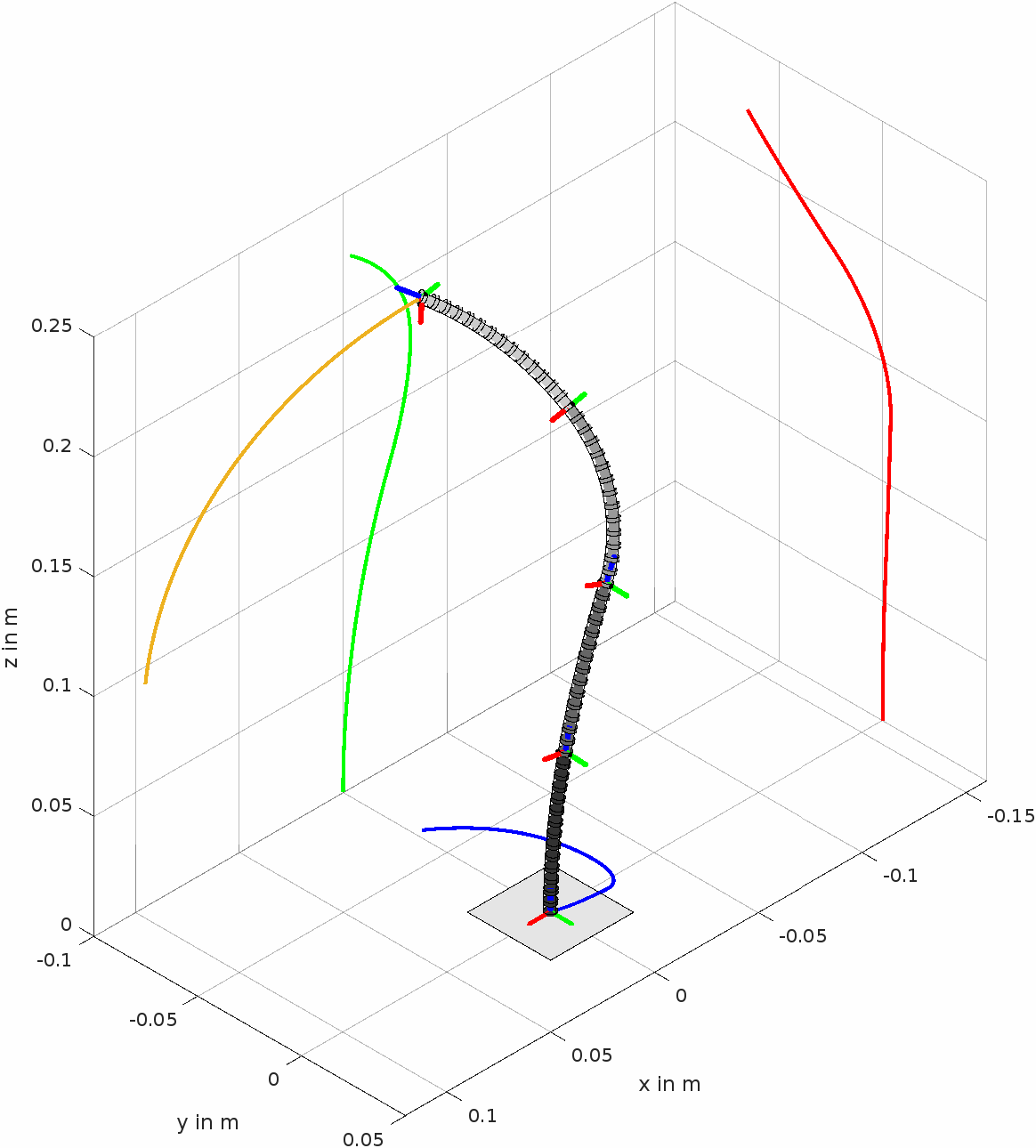}
        \put(0,85){\footnotesize$t = \SI{2.93}{s}$}
    \end{overpic}
    \begin{overpic}[width=0.19\textwidth]{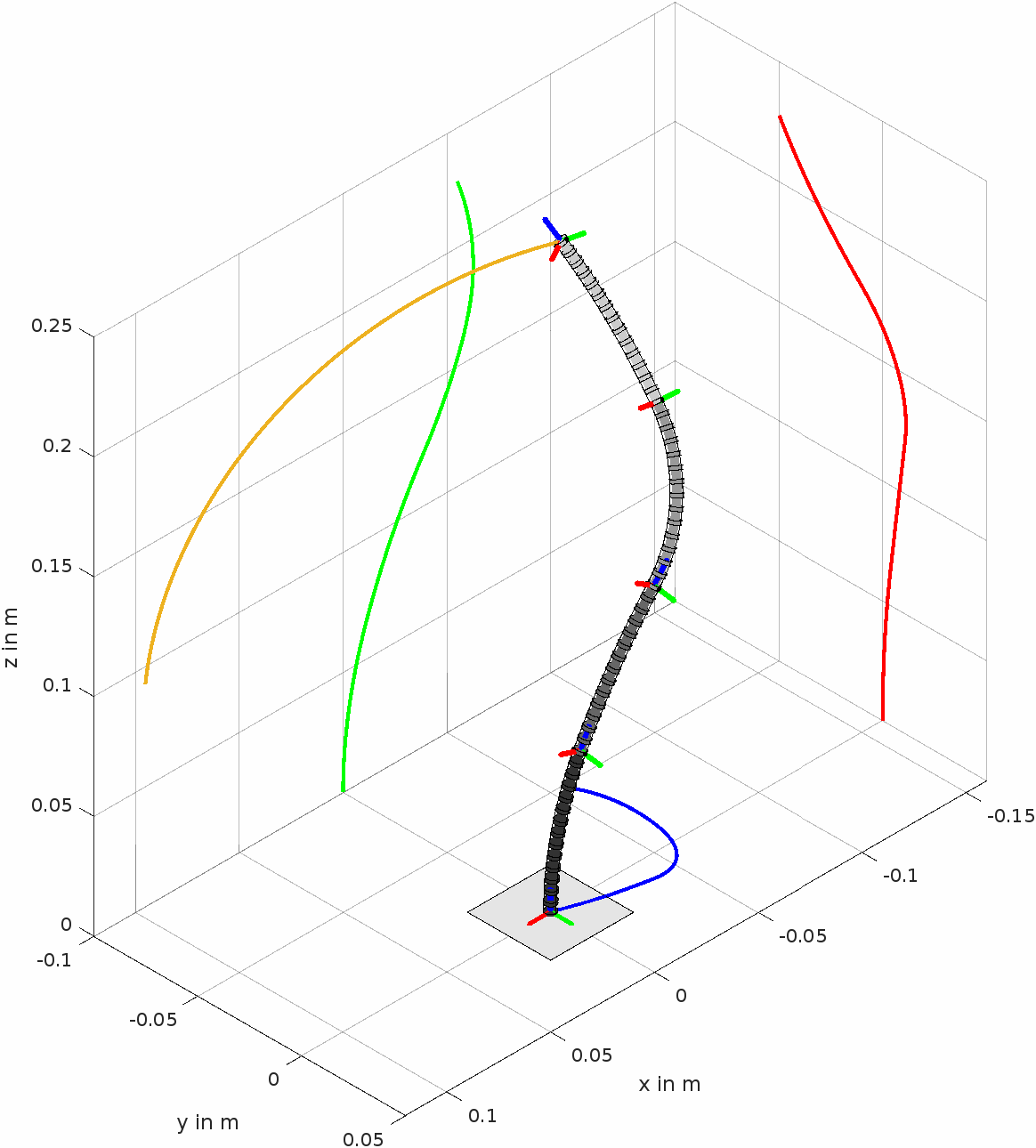}
        \put(0,85){\footnotesize$t = \SI{3.33}{s}$}
    \end{overpic}
    \begin{overpic}[width=0.19\textwidth]{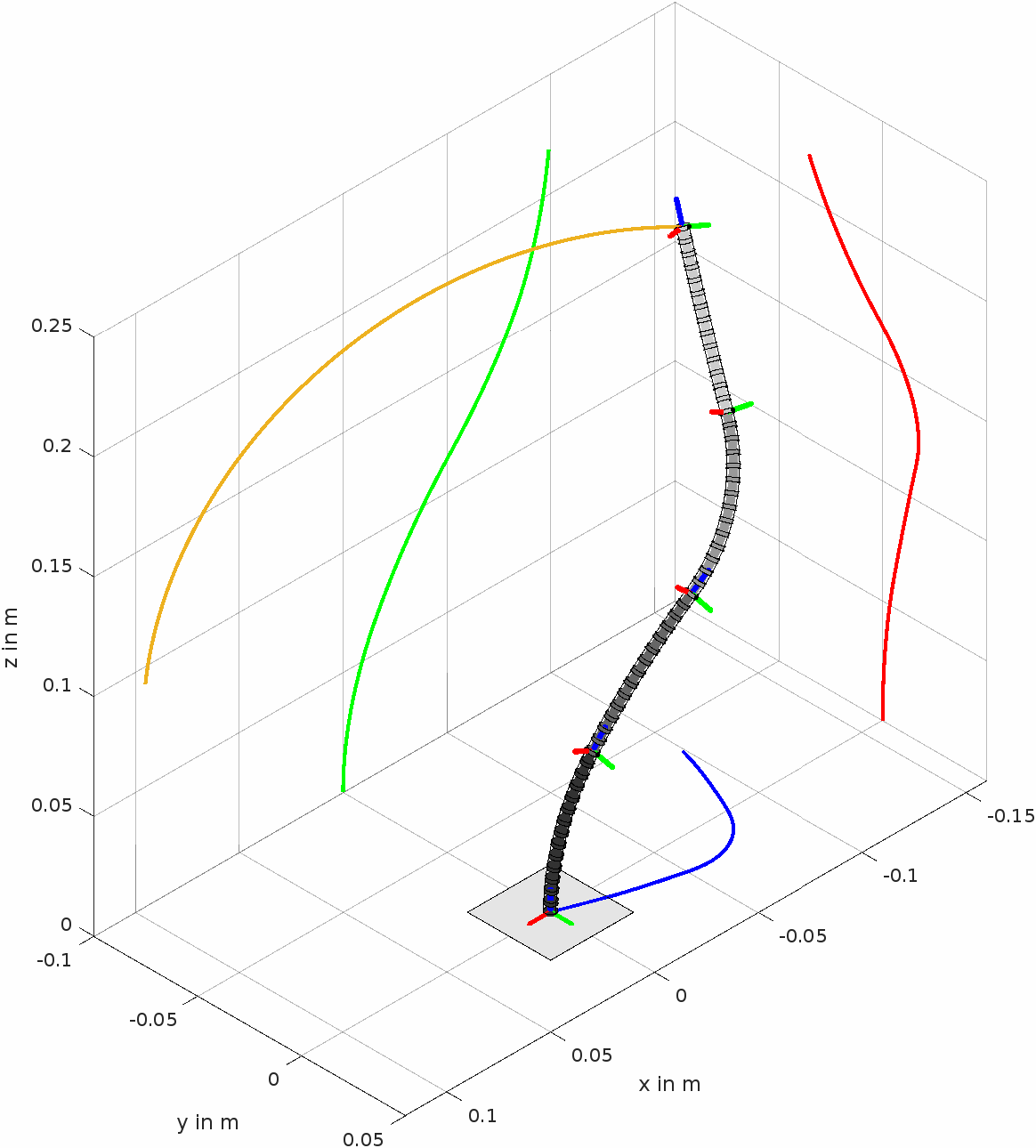}
        \put(0,85){\footnotesize$t = \SI{3.73}{s}$}
    \end{overpic}
    \hfill
    \\[0.25em]
    \hfill
    \begin{overpic}[width=0.19\textwidth]{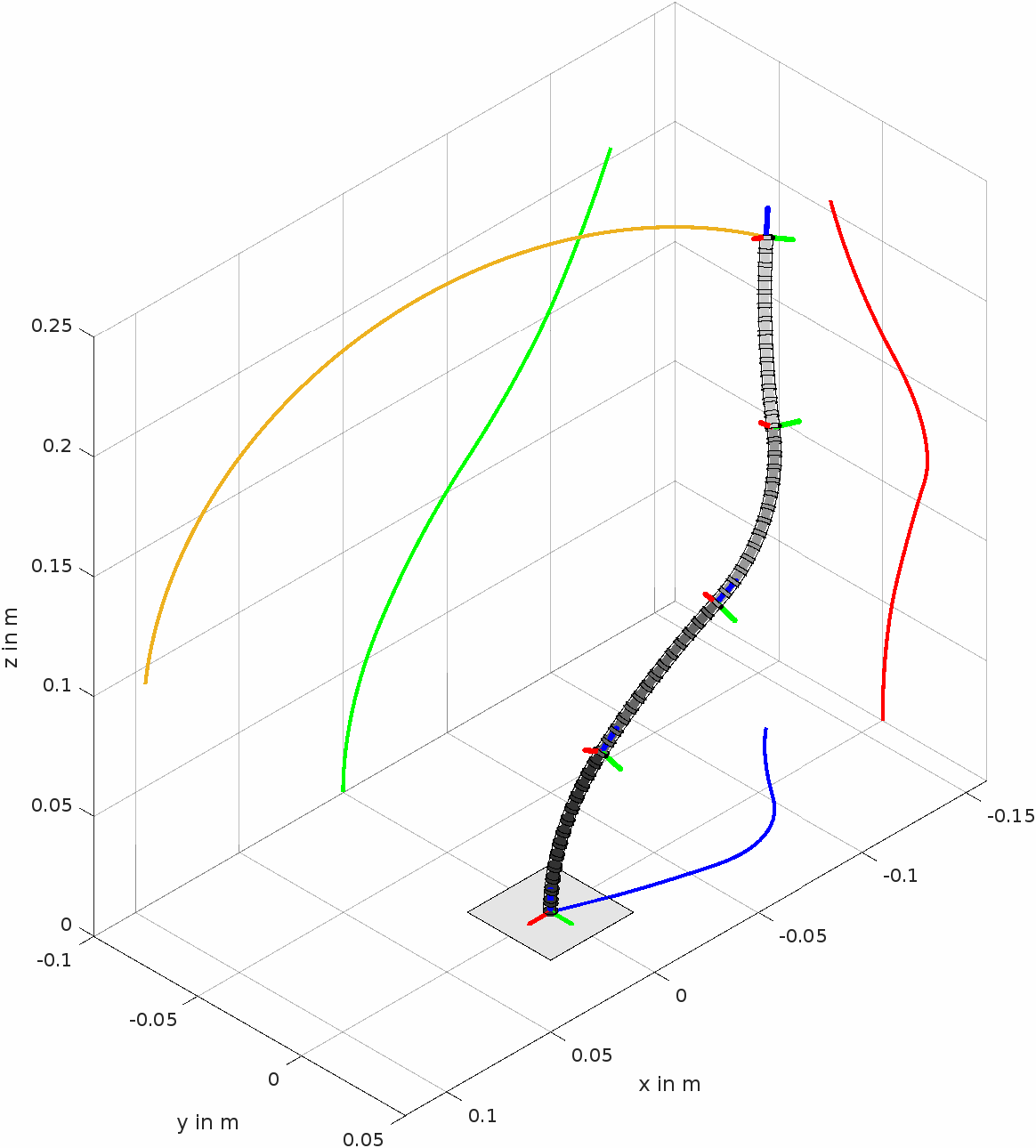}
        \put(0,85){\footnotesize$t = \SI{4.13}{s}$}
    \end{overpic}
    \begin{overpic}[width=0.19\textwidth]{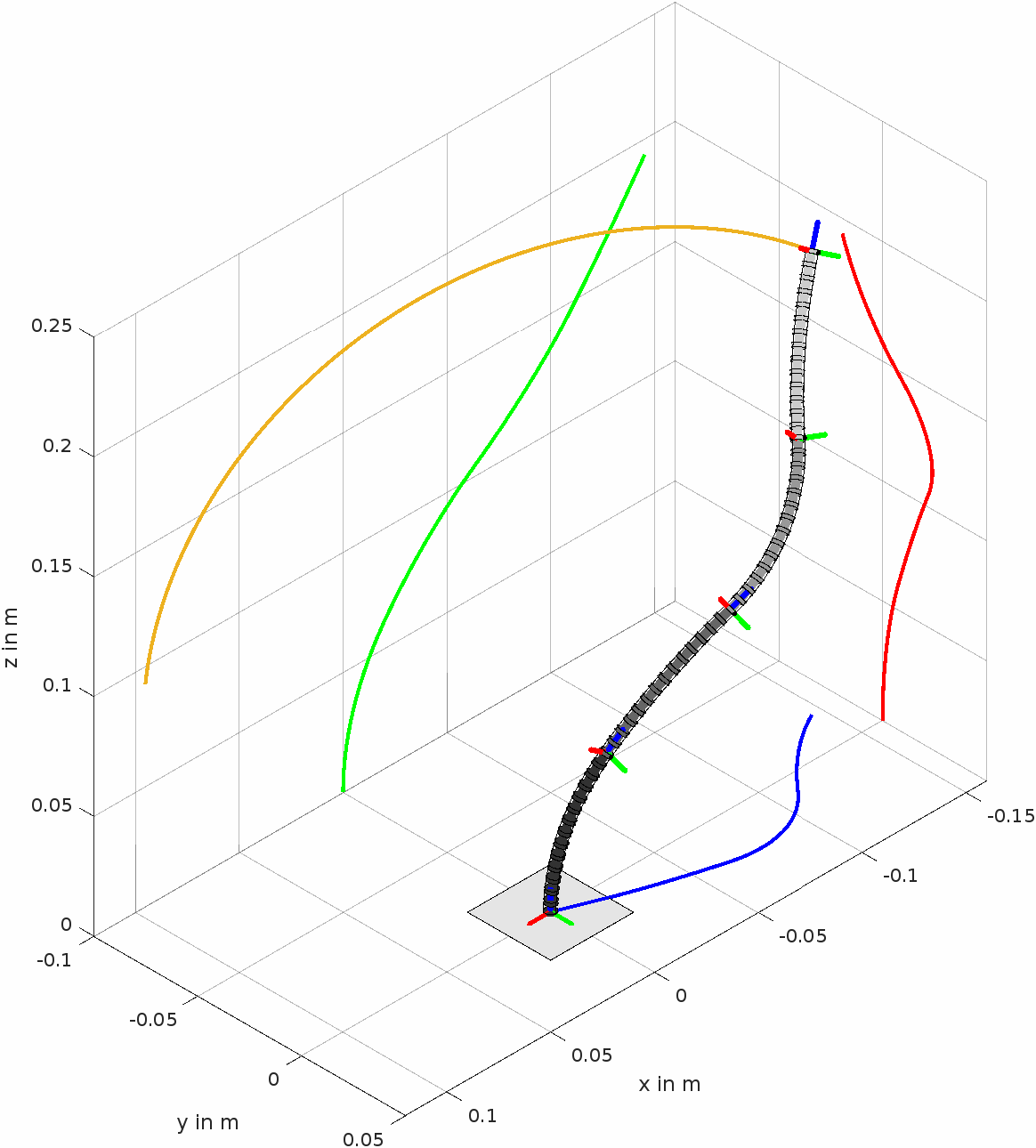}
        \put(0,85){\footnotesize$t = \SI{4.53}{s}$}
    \end{overpic}
    \begin{overpic}[width=0.19\textwidth]{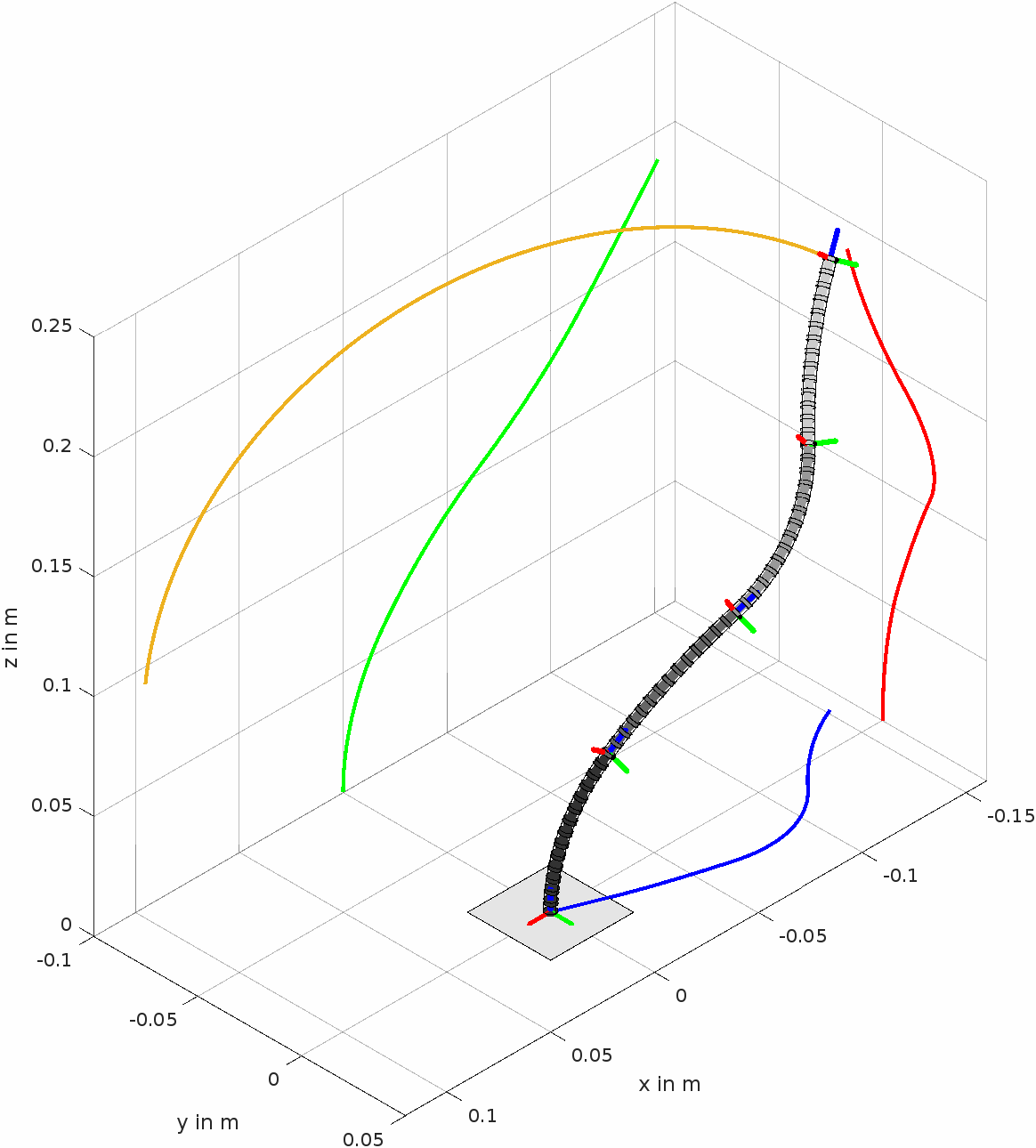}
        \put(0,85){\footnotesize$t = \SI{4.94}{s}$}
    \end{overpic}
    \begin{overpic}[width=0.19\textwidth]{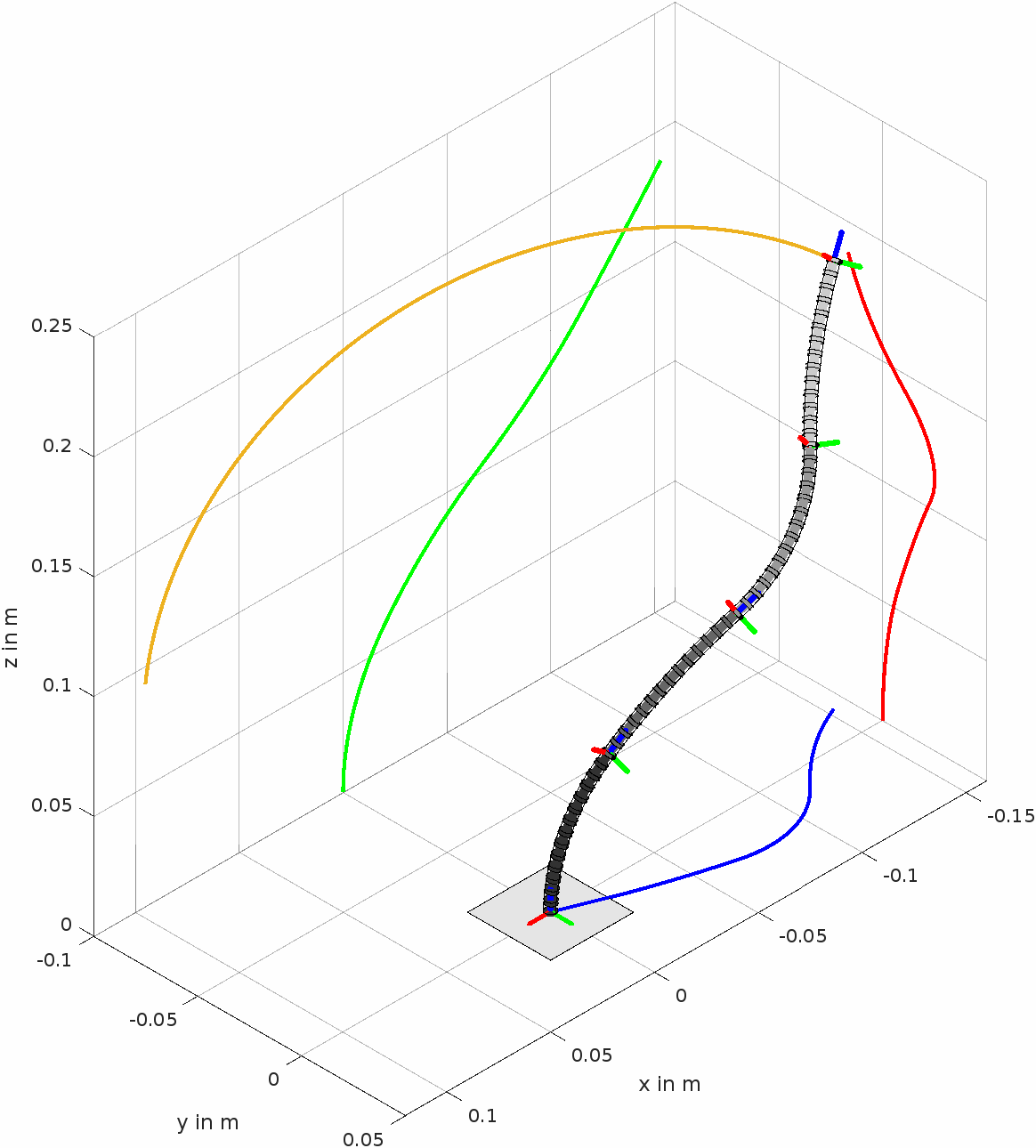}
        \put(0,85){\footnotesize$t = \SI{5.34}{s}$}
    \end{overpic}
    \begin{overpic}[width=0.19\textwidth]{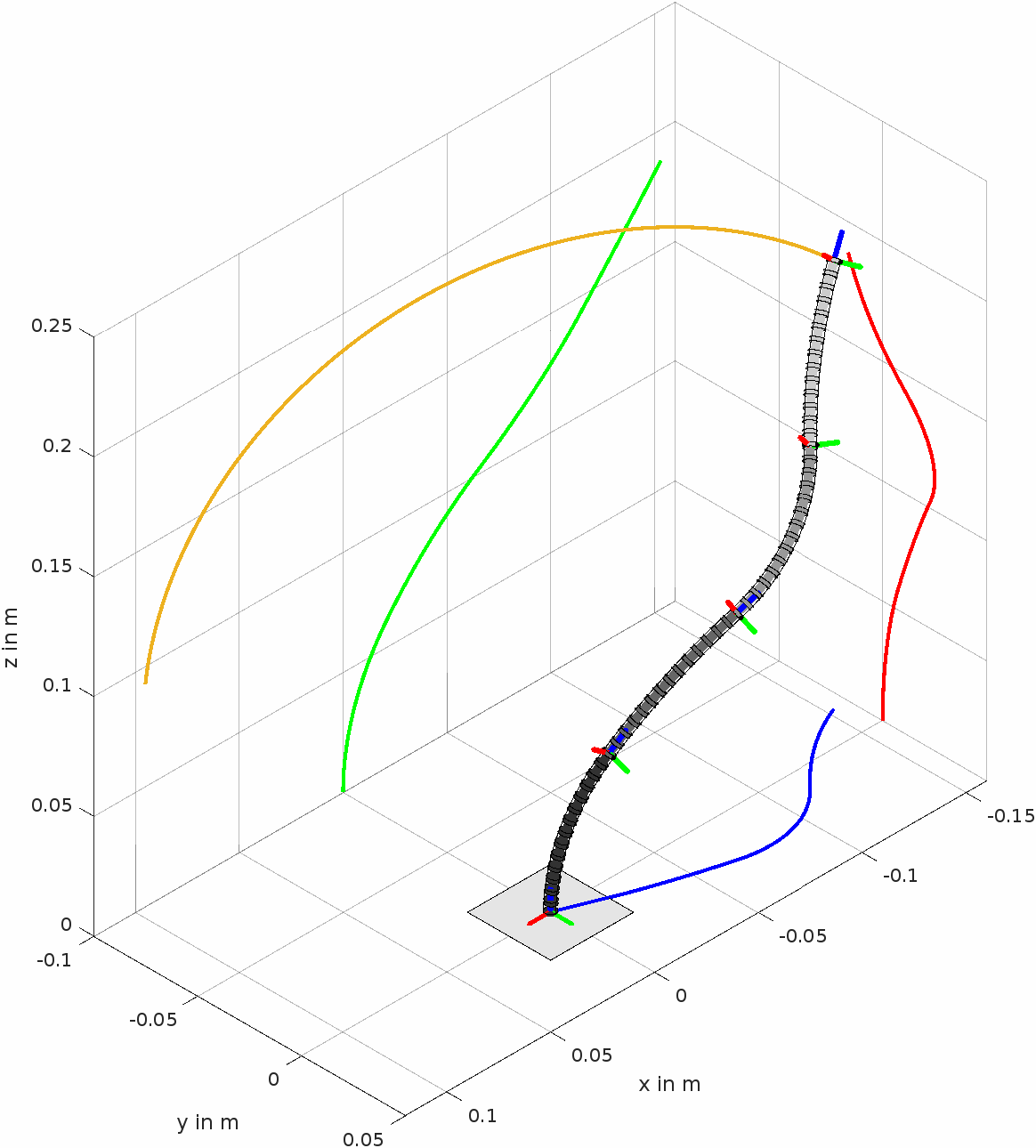}
        \put(0,85){\footnotesize$t = \SI{5.74}{s}$}
    \end{overpic}
    \hfill
    \\[0em]
    \caption{
        Image sequence of a motion.
        The outputs of the controllers are fed to a displacement-actuated continuum robot with four segments.
        For the visualization, the \textsc{CRVisToolkit} is used.
        }
    \label{fig:results_motion}
\end{figure*}

As can be seen from the image sequence in Fig.~\ref{fig:results_motion}, a displacement-actuated continuum robot with four segments can move from a start configuration to a goal configuration.
The course of the desired and executed trajectories are shown in Fig.~\ref{fig:results_closed_loop}.

\section{Discussion and Future Work}

The general approach of the proposed framework is to construct components of the framework using Clarke coordinates.
The components of this framework are a planner, a trajectory generator, and a controller.
Each of them can be switched to a different component utilizing Clarke coordinates.  
Furthermore, using an encoder \eqref{eq:forward}, the framework can bypass components to incorporate approaches that are not based on Clarke coordinates.
Currently, the framework is limited to kinematics, the independence of each segment, and the utilization of relatively simple components.
This can be overcome by incorporating dynamic models and more sophisticated approaches for each component.

Furthermore, the only component with feedback is the controller.
However, a large variety of different approaches with feedback from the robot and environment exist in the literature.
For instance, collision avoidance for static and dynamic environments.
For the former, the formulation of C-space obstacles \cite{Lozano-Perez_TOC_1983} on the manifold is an interesting direction.
For the latter, adapting potential fields \cite{Khatib_IJRR_1986} on the manifold is worth pursuing.
Due to the low dimensionality of the \SI{2}{dof} manifold, it is expected to have the potential to be more computationally efficient.

In general, each of the components is exchangeable.
As shown in Fig.~\ref{fig:framework}, each component has input and output defined as Clarke coordinates.
Moreover, due to the use of Clarke transform \cite{GrassmannSenykBurgner-Kahrs_arXiv_2024}, it inherits the potential benefits of being computationally efficient, interpretable, closed-form, and compact in the formulation. 
One might say that the components are defined in their canonical formulation since Clarke coordinates are the unification and generalization of improved state representations.
However, the cost of time and effort might not justify the reimplementation of an existing framework.
To reap the benefits of particular components, the encoder-decoder architecture \cite{GrassmannBurgner-Kahrs_arXiv_2024} presented in Sec.~\ref{sec:encoder-decoder} allows to include components based on Clarke transform into an existing framework not formulated with Clarke coordinates.
Figure~\ref{fig:controller_encoder-decoder} shows this for the controller component.
This is akin to using quaternions in certain parts of a bigger framework mainly relying on rotation matrices.

Regarding the generation of trajectories, loosely speaking, the trajectory generator \eqref{eq:trajectory} for the joint space is Clarke transformed onto the manifold to form the trajectory generator \eqref{eq:trajectory_clarke} for Clarke coordinates.
While \eqref{eq:trajectory} is transformed using \eqref{eq:MP}, the kinematic constraints for \eqref{eq:trajectory_clarke} are not simply obtained.
Furthermore, the used point-to-point trajectory generation as defined for \eqref{eq:trajectory} and \eqref{eq:trajectory_clarke} are commonly used for independent positions in Task space or independent joint values of a serial kinematic rigid robot.
Therefore, those constraints are not directly related to the joints.
For future work, we will investigate the use of virtual displacement.

To show that the framework is not limited to constant curvature assumption \cite{WebsterJones_IJRR_2010} and to derive necessary relations improving the understanding of sampling and kinematic constraints, we delved into the virtual displacement \cite{FirdausVadali_AIR_2023, GrassmannSenykBurgner-Kahrs_arXiv_2024} and displacement constraint \eqref{eq:sum_rho}.
We establish in Sec.~\ref{sec:virtual_displacement} that the displacement constraint \eqref{eq:sum_rho} is not a result of constant curvature.
Further connections to parallel curves are made that reframe geometric insights by Simaan \textit{et al.} \cite{SimaanTaylorFlint_ICRA_2004} and Burgner-Kahrs \textit{et al.} \cite{Burgner-KahrsRuckerChoset_TRO_2015}.
In the case of displacement-actuated continuum robots, the displacement constraint \eqref{eq:sum_rho} and parallel curves imply that the centerline needs to be sufficiently smooth and the tip orientation $\phi$ is related to displacements.
In contrast and in general, parallel curves for a piecewise linear curve akin to a serial-kinematic robot do not have this property, \textit{i.e.}, \eqref{eq:sum_rho}.

More importantly, the geometric insights on parallel curves imply that no displacement, \textit{i.e.}, $\rhovec = \boldsymbol{0}$, is detected if the tip orientation $\phi$ is the same as the base orientation, \textit{i.e.}, $\phi = 0$.
This is depicted in Fig.~\ref{fig:parallel_curve}.
This has implications for the design of high-level planners, controllers, and sensor placement.
For instance, a proprioceptive sensor based on displacement-actuated joints might not register an interaction with the environment, \textit{e.g.}, unwanted collation, or desired manipulation.
However, non-constant joint location along the structure, \textit{e.g.}, non-straight tendon routing, will create a non-zero displacement constraint, \textit{cf.} \eqref{eq:sum_rho}, in general.
To investigate this challenge, future work will include the extension to more general joint locations varying along the structure of a displacement-actuated continuum robot.
\section{Conclusions}

We lay out a principle to formulate a complete framework using Clarke coordinates.
Besides benefiting from the unified formulation on the \SI{2}{dof} manifold, this framework has two key features.
First, each of the components is interfaced using the Clarke coordinates allowing for modularity that comes with the associated advantages.
Second, an encoder acts as an interface to allow the integration of external components that are not formulated using Clarke coordinates.
Those features allow both in-house and external developers from different labs to integrate, among others, task planners, control schemes, and machine learning algorithms.
This can accelerate innovation and drive the adoption of components that are built for displacement-actuated continuum robots.
To clarify, the present framework does not rely on constant curvature assumption.
We anticipate that this paradigm shift of creating frameworks enables knowledge transfer between research communities to accomplish complex tasks for real-world application with these robot systems.

\addcontentsline{toc}{section}{REFERENCES}
\bibliographystyle{IEEEtran}
\bibliography{IEEEabrv, references}

\end{document}